\documentclass[12pt]{article}

\usepackage{latexsym}

\makeatletter
\def\newleaf{\newpage
\newcount\tmp
\tmp=\c@page
\divide\tmp by 2
\multiply\tmp by 2
\ifnum\c@page=\tmp
~\newpage
\fi
}
\makeatother

\expandafter\ifx\csname proofnewpage\endcsname\relax
\let\proofnewpage=\relax
\fi

\def\color[#1]#2{}

\long\def\nop#1{}

\def\comment{\edef\cps{\the\parskip} \parskip=0.5cm \begingroup \tt}

\expandafter\ifx\csname shortcite\endcsname\relax
\let\shortcite=\cite
\fi

\newbox\current

\long\def\plframebox#1{
\setbox\current\vbox{#1}		

\vbox to \ht\current {\hrule\vss
\hbox to \wd\current {%
\vrule \hss\box\current\hss \vrule}
\vss\hrule }
}

\long\def\eatpar#1{%
\ifx#1\par                      
\let\nextmove=\eatpar           
\else
\let\nextmove=#1
\fi
\noexpand\nextmove
}

\def\modifymargins#1#2{
\newdimen\addtoh
\newdimen\addtow
\addtoh=#1
\addtow=#2

\advance\topmargin by -\addtoh
\multiply\addtoh by 2
\advance\textheight by \addtoh

\advance\oddsidemargin by -\addtow
\advance\evensidemargin by -\addtow
\multiply\addtow by 2
\advance\textwidth by \addtow
}

\begingroup
\catcode`\~=11
\gdef\centertilde#1{\lower #1pt\hbox{~}}
\endgroup

\newcount\currenttime
\newcount\hour
\newcount\minute

\def\printtime{%
\currenttime=\time
\hour=\currenttime
\divide\hour by 60
\minute=-\hour
\multiply\minute by 60
\advance\minute by \currenttime
\the\hour:\ifnum\minute<10 0\fi\the\minute
}

\begingroup
\makeatletter
\global\let\@@date=\@date
\gdef\@date{\@@date\ --- \printtime}
\endgroup

\def\oggi{\number\day\space 
\ifcase\month\or
Gennaio\or Febbraio\or Marzo\or Aprile\or Maggio\or Giugno\or
Luglio\or Agosto\or Settembre\or Ottobre\or Novembre\or Dicembre\fi
\space \number\year}

\newcounter{rmexample}

\def\proof{\noindent {\sl Proof.\ \ }}

\def\qed{\hfill{\boxit{}}
  \ifdim\lastskip<\medskipamount \removelastskip\penalty55\medskip\fi}
\def\qedn#1{\hfill{\boxit{}$_#1$}
  \ifdim\lastskip<\medskipamount \removelastskip\penalty55\medskip\fi}
\long\def\boxit#1{\vbox{\hrule\hbox{\vrule\kern3pt
                  \vbox{\kern3pt#1\kern3pt}\kern3pt\vrule}\hrule}}

  \def\D{{\cal D}}

\def\ie{i.e.}
\def\eg{e.g.}
\def\wrt{w.r.t.}

\def\l{\langle}
\def\r{\rangle}

\def\true{{\sf true}}
\def\false{{\sf false}}

\def\C{{\rm C}}

\def\p{{\rm P}}
\def\np{{\rm NP}}

\def\S#1{\mbox{$\Sigma^p_{#1}$}}
\def\P#1{\mbox{$\Pi^p_{#1}$}}
\def\D#1{\mbox{$\Delta^p_{#1}$}}

\def\pp{{\rm PP}}

\def\nuc#1{\mbox{$\parallel\!\leadsto$#1}}

\def\nucS#1{\nuc{$\Sigma^p_{#1}$}}

\def\profont{\sf}

\def\x3c{{\profont x3c}}

\def\possnewtheorem#1#2{
\expandafter\ifx\csname #1\endcsname\relax
\newtheorem{#1}{#2}
\fi
}

\def\possnewtheoremthree#1[#2]#3{
\expandafter\ifx\csname #1\endcsname\relax
\newtheorem{#1}[#2]{#3}
\fi
}

\possnewtheorem{theorem}{Theorem}
\possnewtheorem{corollary}{Corollary}
\possnewtheorem{lemma}{Lemma}
\possnewtheoremthree{proposition}[theorem]{Proposition}
\possnewtheorem{definition}{Definition}
\possnewtheorem{question}{Question}
\possnewtheorem{example}{Example}
\possnewtheorem{nontheorem}{Counterexample}
\possnewtheorem{property}{Property}
\possnewtheorem{assumption}{Assumption}
\possnewtheorem{conjecture}{Conjecture}
\possnewtheorem{notation}{Notation}
\newenvironment{theorem*}[1]{{\noindent \bf Theorem~#1}\begin{it}}{\end{it}\

}

\def\up{*_{UP}}

\modifymargins{65pt}{40pt}

\def\prec{prec}
\def\cons{cons}
\def\just{just}

\def\proc{proc}

\def\up{{\rm UP}}

\def\mathitem[#1]{
\item[
{\begin{minipage}[b]{0cm}
\[#1\hss\]
\end{minipage}}
]~}

\def\threednf{{\sf 3dnf}}

\let\subsectionnewpage=\newpage

\title{Bijective Faithful Translations among Default Logics}
\author{Paolo Liberatore%
\thanks{Dipartimento di Informatica e Sistemistica,
Universit\`a di Roma La Sapienza,
Via Ariosto 25, 00185, Rome, Italy.
Email: paolo@liberatore.org}
}
\date{}

\begin{document}

\maketitle

\begin{abstract}

In this article, we study translations between variants of
defaults logics such that the extensions of the theories
that are the input and the output of the translation are in
a bijective correspondence. We assume that a translation can
introduce new variables and that the result of translating a
theory can either be produced in time polynomial in the size
of the theory or its output is of size polynomial in the
size of the theory; we restrict to the case in which the
original theory has extensions. This study fills a gap
between two previous pieces of work, one studying bijective
translations among restrictions of default logics, and the
other one studying non-bijective translations between
default logics variants.

\end{abstract}

 %


\section{Introduction}

A translation from one logic to another is {\em faithful} if
it preserves not only the consequences but also the models
of the original theory. What in modal logic is a model, in
default logic \cite{reit-80,besn-89,anto-99} is an
extension; therefore, a faithful translation involving two
default logics is a translation preserving the extensions.

The existence and non-existence of faithful translations
among various logics are known
\cite{imie-87,kono-88,enge-treu-93,gott-95}. Recently, some
effort has been devoted to translations that introduce new
variables
\cite{janh-98,janh-01,janh-03,delg-scha-03,delg-scha-05}:
these translations generate theories which may contain new
variables in addition to the ones of the corresponding
original theories. The addition of new variables allows for
translations that would otherwise be impossible: for
example, no translation that {\em exactly} preserves the
extensions exists from justified default logic to Reiter
default logic; this is because $E_1 \subset E_2$ cannot hold
for two Reiter extensions $E_1$ and $E_2$ of the same
theory, while this situation is instead possible for two
justified extensions \cite{libe-extensions}. Introducing new
variables can however circumvent this difficulty, because
two justified extensions $E_1$ and $E_2$ such that $E_1
\subset E_2$ can be translated into $E_1 \cup E_1'$ and $E_2
\cup E_2'$, respectively, provided that $E_1 \cup E_1'
\not\subset E_2 \cup E_2'$.

The possibility of adding new variables is therefore of
interest because it allows for translations that would
otherwise be impossible. These translations are not defined
in terms of logical equivalence between extensions $E_1
\equiv E_2$, but in terms of var-equivalence $E_1 \equiv_X
E_2$, where $X$ is the set of variables of the original
theory and $E_1 \equiv_X E_2$ means that $E_1$ and $E_2$
have the same consequences when restricted over the alphabet
$X$ \cite{lang-etal-03}.

Faithful translations can be defined in two ways, which are
equivalent when new variables are not allowed. In
particular, a translation is faithful if each theory $T_1$
is translated into a theory $T_2$ such that either:

\begin{enumerate}

\item there is a bijection between the extensions of $T_1$
and the extensions of $T_2$ such that the associated
extensions of $T_1$ and $T_2$ are equivalent, or

\item for every extension of $T_1$ there exists an
equivalent extension of $T_2$ and vice versa.

\end{enumerate}

These two definitions can also be given when new variables
are allowed, by replacing ``equivalence'' with
``var-equivalence''. However, they no longer coincide.
Indeed, the second definition allows a single extension of
$T_1$ to be associated to several extensions of $T_2$. For
example, if $T_1$ is build on variables $\{x\}$ and $T_2$ on
$\{x, y\}$, the second definition allows the same extension
$Cn(\{x\})$ to be associated to the two extensions
$Cn(\{x,y\})$ and $Cn(\{x,\neg y\})$. These two extensions
are indeed var-equivalent to the original one, but are not
classically equivalent to it or to each other. This
translation is faithful according to the second definition
but not according to the first.

This example shows that that the two considered definitions
of faithfulness do not coincide. In this paper, we call
translations satisfying the first definition {\em bijective
faithful} and translations satisfying the second {\em
faithful}. This choice is motivated by the fact that all
translations obeying the first definition also obey the
second but not vice versa, that is, a bijection between the
extensions is an additional requirement over the
translation.

Translations among default theories producing new variables
have been studied by Janhunen~\cite{janh-98,janh-01,janh-03}
and Delgrande and Schaub~\cite{delg-scha-03,delg-scha-05}.
All these authors considered faithful translations, but
using two differing definitions: the former author studies
bijective faithful translations, the latter authors do not
require a bijection between extensions.

In particular, Delgrande and
Schaub~\cite{delg-scha-03,delg-scha-05} have shown faithful
polynomial-time translations from some default logic
variants into Reiter default logic. Some of their faithful
reductions produce a bijection between the extensions only
using a definition of extensions that include the
justification of the applied defaults. In particular, if one
defines an extension to be the deductive closure of the
consequences of the applied defaults, their translation from
justified default logic into Reiter default logic is
bijective, while their translations from constrained and
rational to Reiter default logic are faithful but not
bijective. All of their reductions are bijective if one
takes an extension to be include also the justification of
the applied defaults.

Janhunen~\cite{janh-98,janh-01,janh-03} has instead studied
bijective faithful translations, but not between default
logics variants but between default logics restrictions, and
between default logics and other logics. As a result, the
study of bijective translations between default logics
variants is still largely open, and is the subject of this
article.

The results about the existence of polynomial-time and
polynomial-size bijective faithful translations are shown in
Table~\ref{table}. The existence of polynomial-time
bijective faithful translations from constrained or rational
default logic to Reiter or justified default logic would
have some consequences on complexity classes, whenever
extensions are considered to be the deductive closure of the
consequences of the applied default only (\ie, not including
the justifications). Ideally, a negative result should be
unconditioned (\eg, there is no bijective polynomial time
faithful translations from constrained to Reiter default
logics) or at least conditioned to the collapse of the
polynomial hierarchy (\eg, if there exists a bijective
polynomial-time faithful translations from constrained to
Reiter default logics then the polynomial hierarchy
collapses); unfortunately, none of these two claims could be
proved. We however show some consequences on complexity
classes of the existence of a bijective polynomial time
faithful translations from constrained to Reiter default
logics.

Faithful translations cannot exist from semantics where a
theory may have no extension to semantics where this is not
possible. On the other hand, it can be shown that in some
cases theories having no extensions are {\em the only ones}
that cannot be translated. For example, Reiter default logic
(which allows a theory to have no extensions) cannot be in
general translated into normal default logic (in which every
theory has at least an extension). However, if one restricts
to theories having at least one extension, then Reiter
default logic can be faithfully translated into normal
default logic \cite{libe-failsafe}. Translations that work
in the assumption of existence of extensions are of interest
because theories can be modified in a very simple way so
that they are added a single known extension. Many problems,
such as entailment, number of extensions, etc. can therefore
be solved via such translations.

In this paper, we show bijective faithful translations from
rational and Reiter to constrained default logic and from
Reiter to justified default logic. These translations are
polynomial-time but require not only the original theory to
have an extension, but also that a formula equivalent to one
of the strongest (\ie, minimal \wrt\  set containment)
extension is given. Such translations are of interest
because an extension, being the deductive closure of a set
of consequences of some defaults in the theory, can always
be represented by a polynomially sized formula. The result
of these translations are therefore of size polynomial in
the size of the original theory. In other words, for every
theory in the original semantics (provided it has
extensions) there exists a theory in the resulting semantics
that has size polynomial in that of the original theory.
More concisely, what can be expressed in the first semantics
can also be expressed in the second one in comparable space.
Size-preserving translations of this kind are called
polysize because they produce a result that is polynomial
{\em in size} \wrt\  the size of the input theory. Finally,
we show some consequences of the existence of a bijective
faithful polysize translation from constrained or rational
to Reiter or justified default logic on the counting
hierarchy.

\begin{table}[h]
\begin{minipage}{\textwidth}

\begin{center}
\begin{tabular}{l|c|c|c|c|}
$\downarrow$From~~~~~\hfill To $\rightarrow$
& Reiter	& Justified	& Rational	& Constrained \\
\hline
Reiter	&
yes$^1$ &
no$^2$ &
yes$^3$ &
no$^2$
\\
Justified &
yes$^3$ &
&
yes$^4$ &
yes
\\
Rational &
open$^5$ &
no$^2$ &
&
no$^2$
\\
Constrained &
open$^5$ &
open$^5$ &
yes&

\\
\hline
\end{tabular}

\

Bijective faithful polytime translations between semantics

\vskip 1cm

\begin{tabular}{l|c|c|c|c|}
$\downarrow$From~~~~~\hfill To $\rightarrow$
& Reiter	& Justified	& Rational	& Constrained \\
\hline
Reiter	&
yes$^1$ &
yes &
yes$^3$ &
yes
\\
&
{\footnotesize (polytime)} &
{\footnotesize (strongest extension)} &
{\footnotesize (polytime)} &
{\footnotesize (strongest extension)}
\\
\hline
Justified &
yes$^3$ &
&
yes$^4$ &
yes
\\
&
{\footnotesize (polytime)} &
&
{\footnotesize (polytime)} &
{\footnotesize (polytime)}
\\
\hline
Rational &
open$^5$ &
open$^5$ &
&
yes
\\
& & & &
{\footnotesize (strongest extension)}
\\
\hline
Constrained &
open$^5$ &
open$^5$ &
polytime &

\\
\hline
\end{tabular}

\

Bijective faithful polysize translations between semantics.

\end{center}

\caption{Faithful Translations between Semantics}
\label{table}

\underline{\hbox to 5cm{\hfill}}

\footnotesize

\parindent 20pt

$^1$ Delgrande and Schaub~\cite{delg-scha-03,delg-scha-05}
proved that Reiter default logic can ``simulate itself'',
\ie, there is a non-trivial polynomial time bijective
faithful translation from Reiter default logic to itself.

$^2$ Unless the polynomial hierarchy collapses.

$^3$ Proved by Delgrande and Schaub
\cite{delg-scha-03,delg-scha-05}.

$^4$ Trivially entailed by the reductions from justified to
Reiter default logic and from Reiter to rational default
logic by Delgrande and
Schaub~\cite{delg-scha-03,delg-scha-05}.

$^5$ Effects of the existence of a polysize translation
on the counting hierarchy are shown in this article.

\end{minipage}
\end{table}
 %

\nop{~\newpage ~\newpage}

 %

\section{Definitions}

\let\subsectionnewpage=\relax
\subsection{Default Logics}

We use the operational semantics for default logics. Two
slightly different, but equivalent, operational semantics
for default logics have been given independently by Antoniou
and Sperschneider \cite{anto-sper-94,anto-99} and by
Froidevaux and Mengin~\cite{froi-meng-92,froi-meng-94}. A
default is a rule of the form:

\[
d=\frac{\alpha:\beta}{\gamma}
\]

The formulae $\alpha$, $\beta$, and $\gamma$ are called the
precondition, the justification, and the consequence of $d$,
and are denoted as $\prec(d)$, $\just(d)$, and $\cons(d)$,
respectively. This notation is extended to sets and
sequences of defaults in the obvious way. A default is
applicable if its precondition is true and its justification
is consistent; if this is the case, its consequence should
be considered true.

A default theory is a pair $\l D,W \r$ where $D$ is a set of
defaults and $W$ is a consistent theory, called the
background theory. The assumption that $W$ is consistent is
not standard; however, all known semantics give the same
evaluation when the background theory is inconsistent. We
also make some other assumptions about the default theory:
all formulae are propositional, the alphabet and the set $D$
are finite, and all defaults have a single justification.
The latter assumption is irrelevant for some semantics (such
as constrained default logic) but not for other ones (such
as justified default logic.)

We use semantics of default logics based on sequences of
defaults. We typically denote such sequences by $\Pi$,
$\Pi'$, etc. We also denote by $\Pi \cdot \Pi'$ the sequence
composed of $\Pi$ followed by $\Pi'$. When $\Pi'$ is
composed of a single default $d$, we also denote this
concatenation by $\Pi \cdot d$. Given a sequence $\Pi$ and
one of its defaults $d$, we denote by $\Pi[d]$ the sequence
of defaults preceeding $d$ in $\Pi$. We define a process to
be a sequence of defaults that can be applied starting from
the background theory.

\begin{definition}

A {\em process} of a default theory $\l D,W \r$ is a
sequence of defaults $\Pi$ such that $W \cup \cons(\Pi)$ is
consistent and $W \cup \cons(\Pi[d]) \models \prec(d)$ for
every default $d \in \Pi$.

\end{definition}

The definition of processes only takes into account the
preconditions and the consequences of defaults. This is
because the interpretation of the justifications depends on
the semantics. All semantics select a set of processes that
satisfy two conditions: success and closure.
Intuitively, success means that the justifications of
the applied defaults are not contradicted; closure means
that no other default should be applied.

The particular definitions of success and closure depend on
the specific semantics; in turn, closure can be defined in
terms of applicability of a default. The following are the
definitions used by the variants of default logic considered
in this paper.

\begin{description}

\item[Success:] \

\begin{description}

\item[Local:] for each $d \in \Pi$, the set
$W \cup \cons(\Pi) \cup \{\just(d)\}$ is consistent;

\item[Global:]
$W \cup \cons(\Pi) \cup \just(\Pi)$ is consistent.

\end{description}

\item[Closure:]

\

\begin{description}

\item[Inapplicability:] no default $d \not\in \Pi$
is applicable to $\Pi$; applicability of a default $d$ in
$\Pi$ is defined as $W \cup \cons(\Pi) \models \prec(d)$
and:

\begin{description}

\item[Local Applicability:] $W \cup \cons(\Pi) \cup \just(d)$ is
consistent;

\item[Global Applicability:] $W \cup \cons(\Pi) \cup \just(\Pi) \cup
\just(d)$
is consistent.

\end{description}

\item[Maximality:] for any $d \not\in \Pi$, the sequence
$\Pi \cdot [d]$ is not a successful process.

\end{description}

\end{description}

Reiter default logic uses local success and local
inapplicability closure; justified default logic uses local
success and maximality closure; rational default logic uses
global success and global inapplicability closure;
constrained default logic uses global success and maximality
closure.

The definition of processes by Antoniou and Sperschneider
\cite{anto-sper-94,anto-99} and that by Froidevaux and
Mengin~\cite{froi-meng-92,froi-meng-94} differ mainly in
``when justifications are checked''. In terms of our
definition or processes, Antoniou and Sperschneider do not
allow a sequence of defaults to be a process if the
justification of a default is not consistent with the
background theory and the consequences of the previous
defaults. On the contrary, this is allowed by our definition
and that by Froidevaux and Mengin. To the aim of automated
deduction, the first definition may allow reducing the width
of the tree of processes; on the other hand, the second
definition is slightly simpler from a formal point of view.

The extensions of a default theory can be defined in two
different ways, both based on the set of selected processes.
In this paper, we use the following one: if $\Pi$ is a
successful processes, an extension is $Cn(W \cup
\cons(\Pi))$. This definition is what is actually necessary
for defining query answering: the {\em skeptical}
consequences of a default theory are the formulae that are
entailed by all its extensions; the {\em credulous}
consequences are those implied by some of its extensions.

Extensions for rational and constrained default logic have
been initially defined in a form that is equivalent to the
pair $\l \just(\Pi), Cn(W \cup \cons(\Pi)) \r$, where $\Pi$
is a successful process. This second definition includes in
the extensions also the justification of the applied
defaults. In order to distinguish between extensions
according to the first or to the second definition, we use
different names.

\begin{description}

\item[Extension:] $Cn(W \cup \cons(\Pi))$, where $\Pi$ is a
selected process;

\item[Double Extension:] the pair $\l \just(\Pi), Cn(W \cup
\cons(\Pi) \r$, where $\Pi$ is a selected process.

\end{description}

According to the second definition, two processes composed
of the same defaults in different order always generate two
different extensions. The same is not true for the first
definition of extensions for constrained and rational
default logic. In other words, two processes composed of
different defaults can generate the same extension in these
two semantics, if the extension is defined from consequences
only.

A semantics for default logic is {\em fail-safe} if, for
every default theory, any successful process is the prefix
of a successful and closed process. A successful process of
a fail-safe semantics cannot ``fail'': if we can apply a
sequence of defaults, then an extension will be eventually
generated, possibly after applying some other defaults. In
other words, the situation in which we apply some defaults
but then find out that we do not generate an extension never
occurs. Fail-safeness is a form of commitment to defaults:
if we apply a default, we never end up with contradicting
its assumption.

Fail-safeness can also be seen as a form of monotonicity of
processes \wrt\  to sets of defaults: if a semantics is
fail-safe, then adding some defaults to a theory may only
extend the successful and closed process of the theory and
create new ones. However, this form of monotonicity is not
the same as that typically used in the literature, which is
defined in terms of consequences, not processes. Froidevaux
and Mengin~\cite[Theorem~29]{froi-meng-94} have proved a
result that essentially states that every semantics in which
closure is defined as maximal success is fail-safe.
As a result, justified and constrained default logics are
fail-safe.

We assume that the background theory $W$ is consistent. In
this case, if a semantics is fail-safe, then every default
theory has a successful and closed process: since the
process $[~]$ is successful, a process $\Pi$ that is
successful and closed exists. The condition of
antimonotonicity provides an algorithm for finding this
successful and closed process: if $\Pi$ is successful and
closed, all its initial fragments are successful as well. We
can therefore obtain a successful and closed process by
iteratively adding to $[~]$ a default that leads to a
successful process.

 %

\let\subsectionnewpage=\newpage
\subsection{Translations}

We assume that translations between default logics can
introduce new variables. Technically, this is possible
thanks to the concept of var-equivalence
\cite{lang-etal-03}. In plain terms, two formulae are
var-equivalent if and only if their consequences, when
restricted to be formulae on a given alphabet, are the same.

\begin{definition}[Var-Equivalence]

Two formulae $\alpha$ and $\beta$ are var-equivalent \wrt\
variables $X$ if and only if $\alpha \models \gamma$ iff
$\beta \models \gamma$ for every formula $\gamma$ that only
contains variables in $X$.

\end{definition}

The translations we consider may introduce new variables: a
theory $\l D,W \r$ built on variables $X$ is translated into
a default theory $\l D',W' \r$ built on variables $X \cup
Y$. Faithful translations between default logics based on
var-equivalence of extensions have been considered by
Delgrande and Schaub~\cite{delg-scha-03,delg-scha-05} and by
Janhunen~\cite{janh-98,janh-03}. These authors use slightly
different definitions of faithful translations. Delgrande
and Schaub use the following definition.

\begin{definition}[Faithful Translation]
\label{faithful}

A translation that maps each default theory $\l D,W \r$ into
a default theory $\l D',W' \r$ is {\em faithful} if and only
if each extension of $\l D,W \r$ is var-equivalent \wrt\
the variables of $\l D,W \r$ to at least one extension of
$\l D',W' \r$, and vice versa.

\end{definition}

Equivalently, the set of the extensions of $\l D',W' \r$,
after {\em forgetting} \cite{lang-etal-03} the added
variables, is exactly the same as the set of extensions of
$\l D,W \r$. According to this translation, a single
extension $E$ of $\l D,W \r$ may correspond to several
extensions of $\l D',W' \r$, all var-equivalent to $E$ \wrt\
the variables of $\l D,W \r$.

The translations used by Janhunen \shortcite{janh-03} are
faithful in this sense, but also require a bijection to
exists between the extensions of the two theories.

\begin{definition}[Bijective Faithful Translation]
\label{bijective}

A translation that maps each default theory $\l D,W \r$ into
a default theory $\l D',W' \r$ is {\em bijective faithful}
if and only if each extension of $\l D,W \r$ is
var-equivalent \wrt\  the variables of $\l D,W \r$ to {\em
exactly one} extension of $\l D',W' \r$, and vice versa.

\end{definition}

This definition is only different from the previous one only
because ``is var-equivalent [...] to at least one
extension'' is replaced by  ``is var-equivalent [...] to
exactly one extension''. The second definition requires a
bijection between the sets of extensions to exist. Every
translation that satisfies Definition~\ref{bijective} also
satisfies Definition~\ref{faithful}, but not vice versa.
Note that non-bijective faithful translations implicitly
requires every extension of $\l D',W' \r$ to be associated
with a single extension of $\l D,W \r$ but the converse does
not necessarily hold.

A requirement we impose on the translations is that of being
polynomial. There are two possible definitions of
polynomiality, depending on what is required to be
polynomial: the running time or the size of the produced
output. This difference is important, as some translations
require exponential time but still output a polynomially
large theory. In this paper, we consider three kinds of
translations:

\begin{description}

\item[polynomial:] runs in polynomial time;

\item[strongest extension:] runs in polynomial time but
require one of the strongest extension of the original
theory;

\item[polysize:] produces a polynomially large result.

\end{description}

The existence of a polynomial-time translation from one
semantics to another means that any theory expressed in the
first semantics can be translated in polynomial time into an
equivalent theory in the second one. Such translations are
usually considered good from a computational point of view
because they allow solving problems about the first
semantics using procedures developed for the second one.

However, polynomial-time translations does not tell
everything about the ability of semantics at representing
knowledge. The existence or non-existence of polynomial-time
translations do not give an answer to the question ``is it
true that, for every formula in the first semantics, there
exists a formula in the second semantics that is equivalent
to it and only polynomially larger than it?" A polysize
translation from the first semantics to the second instead
provides a positive answer to this question.

Finally, strongest-extension translations are a particular
kind of polysize translations. They are considered
separately from polysize translations because the time
required by the translation, not only the size of its
produced result, is still bounded by the complexity of
finding one of the strongest extensions of the original
theory. In turn, finding such an extension might be easy in
particular cases such as, for example, when a single
extension is introduced by changing the default theory as
explained below.

\subsection{Theories Having No Extensions}

Semantics for default logics differ as for whether a theory
might or might not have no extension. For example, the
theory $\l \frac{:a}{\neg a}, \emptyset \r$ has no extension
in Reiter and rational default logic. All theories have at
least one extension in justified and constrained default
logics.

This argument has been used to prove that Reiter's default
logics cannot be translated into justified or constrained
default logic by Delgrande and Schaub
\shortcite{delg-scha-03}, and that seminormal default
theories cannot always be translated into normal default
theories by Janhunen \shortcite{janh-03}.

In this paper, we consider translations that work in the
assumption that the theory to be translated has some
extension. The existence of extensions might be guaranteed
because:

\begin{enumerate}

\item the theory encodes a domain in which it is known that
an extension exists; for example, while using default logic
for encoding problem of planning, the particular domain
might guarantee the existence of a plan;

\item theories can be made having extensions by a simple
translations that adds a single known extension to them:

\[
\left\l
\left\{
\frac{:a}{a} ,~
\frac{:\neg a}{\neg a}
\right\}
\cup
\left\{
\left.
\frac{\neg a \wedge \alpha:\beta}{\gamma}
\right|
\frac{\alpha:\beta}{\gamma} \in D
\right\},
W
\right\r
\]

For all considered semantics, this theory has exactly the
extensions of $\l D,W \r$ with $\neg a$ added to them plus
the single extension $Cn(a)$. This proves that a very simple
change can make theories guaranteed to have extensions. The
resulting theory only have an easy recognizable added
extension. Querying this theory produces results that are
either identical to those of the original theory (\eg,
entailment) or similar (counting the extensions).

\end{enumerate}

This point is relevant to the present article because we
prove that some translations are possible if the default
theory to translate is assumed to have extensions. In
particular, we show translations from rational to
constrained default logic, and from Reiter to constrained
and justified default logic; such translations would be
impossible if the theory to be translated lacks extensions.
We can therefore conclude that the possible lack of
extensions {\em is the only reason} why such a translation
is impossible in general.

Since these translations are polysize, checking the
existence of extensions does not introduce an additional
cost (in terms of size). The translation works even in the
case of no extensions by slightly extending the syntax and
semantics so that a default theory is either a pair $\l D,W
\r$ or the special symbol $\bot$, which is not assigned any
extensions by the semantics. Thanks to this minimal change,
the translation from, for example, Reiter to justified
default logic can be extended to theories having no
extensions by translating such theories into $\bot$.

A similar change can be done without changing the syntax or
semantics but adding an exception to the definition of
faithfulness, so that a theory with no extensions can be
translated into a theory with a single inconsistent
extension. This way, we can translate theories not having
extensions into $\l \emptyset, \{\bot\} \r$. This
translation preserve the number of extensions unless the
original theory have none, and also preserve the skeptical
consequences exactly.

 %

 %

\section{Polynomial Time Translations}

In this section we show two polynomial-time bijective
faithful translations, one from constrained to rational
default logics, the other from justified to constrained
default logics. We also show that the existence of such
translations from either Reiter or rational default logic to
either justified or constrained default logic implies that
$\S{2} \subseteq \P{2}$. We also prove that the existence of
such translations from either constrained or rational
default logic to either Reiter or justified default logic
implies that $\np^\np = \up^\np$.

\let\subsectionnewpage=\relax
\subsection{From Constrained to Rational}

Constrained default logic can be translated into rational
default logic by simply making all defaults seminormal. In
other words, a theory $\l D,W \r$ is translated into the
following one.

\begin{eqnarray*}
T_{CR}\left(\frac{\alpha:\beta}{\gamma}\right)
&=&
\frac{\alpha:\beta \wedge \gamma}{\gamma}
\\
T_{CR}(\l D,W \r)
&=&
\l \{ T_{CR}(d) ~|~ d \in D \}, W \r
\end{eqnarray*}

We prove that the processes of $\l D,W \r$ are
translated into the processes of $T_{CR}(\l D,W
\r)$.

\begin{lemma}

There exists a bijection between the constrained processes
of $\l D,W \r$ and the rational processes of $T_{CR}(\l D,W
\r)$ such that the extensions generated by two associated
processes are equivalent.

\end{lemma}

\proof We define the translation of a process
$[d_1,\ldots,d_n]$ as $T_{CR}([d_1,\ldots,d_n]) = [
T_{CR}(d_1), \ldots, T_{CR}(d_n) ]$. Let $\Pi$ be a
constrained process of $\l D,W \r$. We show that
$\Pi'=T_{CR}(\Pi)$ is a rational process of $T_{CR}(\l D,W
\r)$.

\begin{description}

\item[$\Pi'$ is a process.] The fact that $W \cup
\cons(\Pi'[d]) \models \prec(d')$ for every $d' \in \Pi'$
follows from the fact that the same condition holds for the
original process $\Pi$, and preconditions and consequences
are not changed by the translation.

\item[$\Pi'$ is globally successful.] This condition holds
because $\just(\Pi') \cup \cons(\Pi') = \just(\Pi) \cup
\cons(\Pi)$, and $\Pi$ is successful.

\item[$\Pi'$ is closed.] We have to prove that no default is
applicable to $\Pi'$. Since $\Pi$ is maximally globally
successful, either $W \cup \cons(\Pi) \not\models \prec(d)$
or $\Pi \cdot [d]$ is not globally successful. In the first
case, $d'=T_{CR}(d)$ is not rationally applicable to $\Pi'$.
In the second case, we have that $W \cup \just(\Pi \cdot
[d]) \cup \cons(\Pi \cdot [d])$ is not consistent. By
definition, $\just(\Pi' \cdot [d']) \cup \cons(\Pi') =
\just(\Pi \cdot [d]) \cup \cons(\Pi \cdot [d])$ because the
translation adds the consequence of each default to its
justification. As a result, we have that $W \cup \just(\Pi')
\cup \cons(\Pi') \cup \just(d')$ is inconsistent; therefore,
$d'$ cannot be applied to $\Pi'$.

\end{description}

We now show the converse: if $\Pi' = T_{CR}(\Pi)$ is a
rational process of $T_{CR}(\l D,W \r)$, then $\Pi$ is a
constrained process of $\l D,W \r$. As before, we denote by
$d'$ the result of translating the single default $d$.

\begin{description}

\item[$\Pi$ is a process.] As in the previous case, since
preconditions and consequences are not changed by the
reduction, if $\Pi'$ is a process so is $\Pi$.

\item[$\Pi$ is successful.] $W \cup \cons(\Pi) \cup
\just(\Pi)$ is consistent because it is equal to $W \cup
\cons(\Pi') \cup \just(\Pi')$, which is consistent because
$\Pi'$ is a globally successful process.

\item[$\Pi$ is maximally successful.] We have to prove that,
if $W \cup \cons(\Pi) \models \prec(d)$, then $\Pi \cdot
[d]$ is not a successful process. Since $\Pi'$ is closed
according to global applicability, if $W \cup \cons(\Pi')
\models \prec(d')$, then $W \cup \cons(\Pi') \cup \just(\Pi)
\cup \just(d')$ is inconsistent. Since the latter is equal
to $W \cup \cons(\Pi') \cup \just(\Pi) \cup \cons(d') \cup
\just(d') = W \cup \cons(\Pi \cdot [d]) \cup \just(\Pi \cdot
[d])$, its inconsistency implies that $\Pi \cdot [d]$ is not
maximally successful.

\end{description}
\qed

The following corollary easily follows.

\begin{corollary}

The constrained extensions of $\l D,W \r$ and the rational
extensions of $T_{CR}(\l D,W \r)$ are the same.

\end{corollary}

In this very simple case we were able to show a faithful
translation that does not introduce new variables, but this
is not generally possible. Remarkably, beside the empty set
of extensions, constrained and rational default logic are
able to express exactly the same sets of extensions
\cite{libe-extensions}.

 %

\let\subsectionnewpage=\newpage
\subsection{From Justified to Constrained}

The semantics of justified default logic is based on a local
consistency check, in which each justification is checked
against the combined consequences of all defaults in the
process. This kind of consistency check can be simulated in
constrained default logic by using a separate alphabet for
each justification. Assume that the original theory contains
$m$ defaults $d_1,\ldots,d_m$ and its variables are those
in a set $X$. The $i$-th default is translated as follows.

\[
T_{JC}
\left(
\frac{\alpha:\beta}{\gamma}, i
\right)
=
\frac{\alpha:\beta[X/X_i]}
{\gamma \wedge \gamma[X/X_1] \wedge \cdots \wedge \gamma[X/X_m]}
\]

This translation assumes a total ordering over the defaults
of the original theory. Processes are translated in the
obvious way, while a default theory is translated as
follows.

\[
T_{JC}( \l \{d_1,\ldots,d_m\},W \r
=
\l
\{T_{JC}(d_1,1), \ldots, T_{JC}(d_m,m)\}
,~
W \wedge W[X/X_1] \wedge \cdots \wedge W[X/X_m]
\r
\]

For every default $d_i$ we have an alphabet $X_i$. The
justification of each default $d_i$ is translated into its
associated alphabet $X_i$. Whenever a default is applied its
consequence $\gamma$ is drawn on all alphabets, so that each
justification is checked separately.

\begin{lemma}

There exists a bijection between the justified processes of
$\l D,W \r$ and the constrained processes of $T_{JC}(\l D,W
\r)$ such that the extensions generated by two associated
processes are var-equivalent \wrt\  the variables of $\l D,W
\r$.

\end{lemma}

\proof We show a correspondence between each justified
process $\Pi$ of $\l D,W \r$ and its corresponding process
$\Pi'=T_{JC}(\Pi)$ of $T_{JC}(\l D,W \r)$. By definition, $W
\cup \cons(\Pi)$ is var-equivalent to $W \wedge W[X/X_1]
\wedge \cdots \wedge W[X/X_m] \cup \cons(\Pi')$ because the
background theory and the consequences of translated
defaults contain the corresponding formulae of the original
one and other ones that do not affect the value of the
variables $X$.

We prove that $\Pi$ is a justified process of the original
theory if and only if $\Pi'=T_{JC}(\Pi)$ is a constrained
process of the translated theory. The following sequence of
equations relates $\Pi$ and $\Pi'$:

\begin{eqnarray*}
\lefteqn{W' \cup \cons(\Pi') \cup \just(\Pi') = } \\
&=&
W \cup
\bigcup_{i=1,\ldots,m} W[X/X_i] \cup
\bigcup_{d_i' \in \Pi'} \just(d_i') \cup
\bigcup_{d_i' \in \Pi'} \cons(d_i') \\
&=&
W \cup
\bigcup_{d_i \not\in \Pi} W[X/X_i] \cup
\bigcup_{d_i \in \Pi} W[X/X_i] \cup \\
&& \cup
\bigcup_{d_i \in \Pi} \just(d_i)[X/X_i] \cup
\bigcup_{d_i \in \Pi} \cons(d_i) \cup
\bigcup_{d_j \in \Pi \atop i=1,\ldots,m} \cons(d_j)[X/X_i] \\
&=&
W \cup
\bigcup_{d_i \in \Pi} \cons(d_i) \cup \\
&& \cup 
\bigcup_{d_i \not\in \Pi} W[X/X_i] \cup 
\bigcup_{d_j \in \Pi \atop d_i \not\in \Pi} \cons(d_j)[X/X_i] \\
&& \cup
\bigcup_{d_i \in \Pi}
\left(
W[X/X_i] \cup
\just(d_i)[X/X_i] \cup
\bigcup_{d_j \in \Pi} \cons(d_j)[X/X_i]
\right) \\
&=&
W \cup
\bigcup_{d_i \in \Pi} \cons(d_i) \cup \\
&& \cup
\bigcup_{d_i \not\in \Pi}
\left(
W \cup
\bigcup_{d_j \in \Pi} \cons(d_j)
\right)[X/X_i] \\
&& \cup
\bigcup_{d_i \in \Pi}
\left(
W \cup
\just(d_i) \cup
\bigcup_{d_j \in \Pi} \cons(d_j)
\right)[X/X_i] \\
\end{eqnarray*}

The consistency of such formula is equivalent to the
consistency of all formulae $W \cup \just(d_i) \cup
\bigcup_{d_i \in \Pi} \cons(d_i)$ for every $d_i \in \Pi$,
because all these formulae are on different alphabets and
their consistency entails the consistency of $W \cup
\bigcup_{d_i \in \Pi} cons(d_i)$ and of all its variants on
the alphabets $X_i$ for $d_i \not\in \Pi$. The proof of the
lemma is based on this fact: the global successfulness of
$\Pi'$ is equivalent to the local successfulness of $\Pi$.

Let us first assume that $\Pi$ is a justified process of $\l
D,W \r$, and show that $\Pi'$ is a constrained process of
$T_{JC}(\l D,W \r)$.

\begin{description}

\item[$\Pi'$ is a process.] This is because the precondition
of the defaults are not changed by the translation and the
background theory and the consequence of each translated
default $d'$ are var-equivalent to the consequence of the
original default $d$ on the variables $X$.

\item[$\Pi'$ is globally successful.] This fact holds
because the global successfulness of $\Pi'$ is equivalent to
the local successfulness of $\Pi$, as shown above.

\item[$\Pi'$ is maximally globally successful.] We have to
prove that, for every default $d_i'$ such that $W' \cup
\cons(\Pi') \models \prec(d_i')$, the formula $W' \cup
\cons(\Pi' \cdot [d_i']) \cup \just(\Pi' \cdot [d_i'])$ is
inconsistent. Let us therefore assume that $W' \cup
\cons(\Pi') \models \prec(d_i')$, which implies that $W \cup
\cons(\Pi) \models \prec(d_i)$. Since $\Pi$ is a maximal
locally successful process, we have that $\Pi \cdot [d_i]$
is not locally successful. As a result, $\Pi' \cdot [d_i']$
is not globally successful.

\end{description}

Let us now assume that $\Pi'$ is a constrained process of
$T_{JC}(\l D,W \r)$, and prove that $\Pi$ is a justified
process of $\l D,W \r$.

\begin{description}

\item[$\Pi$ is a process.] As in the previous case, since
$\Pi'$ is a process, we have that $W' \cup \cons(\Pi'[d'])
\models \prec(d')$. Since $\prec(d')=\prec(d)$ and $W' \cup
\cons(\Pi'[d'])$ is var-equivalent to $W \cup \cons(\Pi)$ on
the variables $X$, we have that $W \cup \cons(\Pi) \models
\prec(d)$.

\item[$\Pi$ is locally successful.] This is because $\Pi'$
is globally successful, and this condition implies the local
success of $\Pi'$.

\item[$\Pi$ is maximally locally successful.] We have to
show that, if $W \cup \cons(\Pi) \models \prec(d_i)$, then
$\Pi \cdot [d_i]$ is not successful. The condition that $W
\cup \cons(\Pi) \models \prec(d_i)$ implies that $W' \cup
\cons(\Pi') \models \prec(d_i')$. As a result, $\Pi' \cdot
[d_i']$ is not globally successful. As a result $\Pi \cdot
[d_i]$ is not locally successful.

\end{description}

We have therefore proved that the justified processes of the
original theory correspond to the constrained processes of
the translated theory. The var-equivalence is guaranteed by
the fact that the alphabets are disjoint.~\qed

This lemma allows proving that the translation from
justified to constrained default logic is faithful.

\begin{corollary}

There exists a bijection from the justified extensions of
$\l D,W \r$ to the constrained extensions of $T_{JC}(\l D,W
\r)$ such that two associated extensions are var-equivalent
\wrt\  the variables of $\l D,W \r$.

\end{corollary}

 %

\subsection{From Reiter or Rational to Justified or Constrained}

Regarding translations from Reiter or rational default logic
into justified or constrained default logic, of course the
translation is in general impossible, as the first two
semantics might not have extensions while the latter always
has. We consider the specific case in which theories are
known to have extensions.

We show that no bijective-faithful polynomial-time
translation from rational default logic to any failsafe
semantics (such as constrained or justified default logic)
exists even if the original theory is known to have
extensions, unless the polynomial hierarchy collapses. The
same claim has been proved for Reiter semantics in another
paper \cite{libe-failsafe}.

This claim is proved by showing that the problem of
entailment in theories having a single extension is $\S{2}
\cap \P{2}$-hard for the rational semantics. The same
problem is in \D{2}\  for every failsafe semantics because
it amounts to generating a single extension, and this
generation is in \D{2}\  because it can be done by applying
one of the applicable defaults and iterating.

\begin{lemma}

The problem of deciding the existence of rational extensions
of a theory having at most one rational extension and an
empty background theory is $\S{2}$-hard.

\end{lemma}

\proof Given $\exists X \forall Y.F$, we build the
following theory.

\[
\left\l
\left\{
\frac{:x_iz_i}{z_i \wedge (a \rightarrow x_i)}, ~
\frac{:\neg x_iz_i}{z_i \wedge (a \rightarrow \neg x_i)}
\right\}
\cup
\left\{
\frac{z_1 \ldots z_n \wedge (a \rightarrow F):}{\neg a} ,~
\frac{z_1 \ldots z_n : a}{\false}
\right\}
,
\emptyset
\right\r
\]

The two defaults corresponding to the variable $x_i$ have
mutually inconsistent justifications $x_iz_i$ and $\neg
x_iz_i$. Once one of them is applied, its justification
disallows the application of the other one.

Since no other default can be applied until all $z_i$'s are
derived, a process is not closed until either one of the two
defaults associated to $x_i$ is applied. What results from
this application is a truth interpretation over the
variables $X$ conditioned to the variable $a$.

If this interpretation entails $F$ regardless of the value
of the variables $Y$, we can apply the default having $\neg
a$ as a conclusion. The result of this application is that
of making all formulae $a \rightarrow x_i$ and $a
\rightarrow \neg x_i$ derived so far vacuous, and the last
default not applicable.

On the other hand, unless $a$ is derived by the application
of this default, the last default can be applied generating
a failure. This means that the theory has at most one
extension, and that happens exactly when $\exists X \forall
Y . F$.~\qed

This lemma can be used for deriving the complexity of the
problem of entailment for theories having a single
extension.

\begin{theorem}

The problem of entailment is $\S{2} \cap \P{2}$-hard for
rational default logic even if the default theory is
guaranteed to have exactly one extension.

\end{theorem}

\proof We have shown that every problem in \S{2}\  can be
reduced to checking whether a theory having zero or one
extension has in fact one extension in rational default
logic. Now, consider that for every instance of a problem in
$\S{2} \cap \P{2}$ we can produce two theories, the first
having 0/1 extensions, and the second having 1/0 extensions,
depending on the instance of the problem. We can then use a
single variable separating the two cases: we combine these
two theories $\l D,\emptyset \r$ and $\l D',\emptyset \r$ by
introducing a new variable $b$, which is either true or
false in each extension of the following default theory:

\[
\left\l
\left\{
\frac{:b}{b} ~,~
\frac{:\neg b}{\neg b}
\right\}
\cup
\left\{
\left.
\frac{b \wedge \alpha:\beta}{\gamma}
\right|
\frac{\alpha:\beta}{\gamma} \in D
\right\}
\cup
\left\{
\left.
\frac{\neg b \wedge \alpha:\beta}{\gamma}
\right|
\frac{\alpha:\beta}{\gamma} \in D'
\right\}
,
\emptyset
\right\r
\]

This theory has all extensions of the two original theories.
In this case, it has a single extension, implying something
or something else depending on the original instance.~\qed

What has been proved so far using complexity classes is that
there is no poly-time reduction from rational default logic
to any fail-safe semantics.

\begin{corollary}

If there exists a bijective faithful polynomial-time
translation from either Reiter or rational default logic to
either justified or constrained default logics then $\S{2}
\subseteq \P{2}$.

\end{corollary}

 %

\subsection{From Constrained and Rational to Reiter and Justified}

Delgrande and Schaub~\cite{delg-scha-03,delg-scha-05} showed
polynomial-time faithful translations from all four
considered semantics to Reiter default logic. The
translations from constrained and rational default logic are
bijective only when double extensions are used. This is
because these two translations copy the justifications of
defaults into their consequence. As a result, two
constrained or rational processes generating the same
extension correspond to two different Reiter extensions. The
correspondence is instead a bijection when double extensions
are used instead. Consider the following theory in either
constrained or rational default logic.

\[
\left\l
\left\{
\frac{:a}{b} ,~
\frac{:\neg a}{b}
\right\}
,~
\emptyset
\right\r
\]

Constrained default logic selects two processes, each one
composed of a single default. These two processes generate
the same extension $Cn(b)$, but two different double
extensions $\l a, Cn(b) \r$ and $\l \neg a, Cn(b) \r$. This
theory is translated into  the following theory in Reiter
default logic:

\[
\left\l
\left\{
	\frac{:a' \wedge b'}{b \wedge a' \wedge b'},
	\frac{:\neg a' \wedge b'}{b \wedge \neg a' \wedge b'}
\right\},
\emptyset
\right\r
\]

The Reiter processes of this theory are still composed of
one default each. However, these processes generate not only
two different double extensions $\l a' \wedge b', Cn(b
\wedge a' \wedge b') \r$ and $\l \neg a' \wedge b', Cn(b
\wedge \neg a' \wedge b') \r$, but also two different
extensions $Cn(b \wedge a' \wedge b')$ and $Cn(b \wedge \neg
a' \wedge b')$. This translation is therefore bijective on
double extensions but not on extensions, as a single
extension of the original theory corresponds to two
extensions of the resulting theory.

An open question is therefore whether bijective faithful
polynomial-time reductions from constrained and rational
default logic to Reiter exist, when extensions, rather than
double extensions, are considered. We show the effects of
existence of such translations on complexity classes.

Checking whether a formula is equivalent to an extension of
a default theory in the constrained or rational semantics is
\S{2} complete \cite{libe-extensions}. The following lemma
proves that the problem remains hard even if the theory is
known to have either one or two extensions, both known in
advance.

\begin{lemma}
\label{one-or-two}

For any formula $F$ over variables $X \cup Y$ one can build
in polynomial time a default theory $\l D,W \r$ whose
rational and constrained extensions are $Cn(\neg a \wedge
\neg b)$ and, if $\exists X \forall Y . F$ is valid,
$Cn(\neg a \wedge b)$.

\end{lemma}

\proof Let $X=\{x_1,\ldots,x_n\}$. The default theory
corresponding to $\exists X \forall Y . F$ is $\l
D,\emptyset \r$ where:

\[
D=
\left\{
\left.
\frac{:x_i}{a \rightarrow x_i} ,~ 
\frac{:\neg x_i}{a \rightarrow \neg x_i} 
\right|
1 \leq i \leq n
\right\}
\cup
\left\{
\frac{a \rightarrow F:\neg a b}{\neg a b} ,~
\frac{:\neg a \neg b}{\neg a \neg b}
\right\}
\]

Applying the last default generates $\neg a \neg b$, which
makes the second-last default inapplicable and entails all
consequences of all other default. The formula $\neg a \neg
b$ is therefore always an extension of this theory.

The application of the defaults in the first set in either
the rational or the constrained semantics lead to a partial
truth evaluation over the variables $x_i$ conditioned to
$a$. We can then apply the second-last default and generate
$b$ if and only if $F$ is valid for this partial truth
evaluation. As a result, this theory has always the
extension $Cn(\neg a \wedge \neg b)$, and also has the
extension $Cn(\neg a \wedge b)$ if and only if $\exists X
\forall Y . F$ is valid.~\qed

The complexity of some problems easily follow from this
lemma.

\begin{corollary}

Checking whether $E$ is a rational or constrained extension
of a default theory is \S{2}-hard, and this result holds
even if the theory has either one or two extensions.

\end{corollary}

\begin{corollary}

Checking whether a default theory has at least two
constrained or rational extensions is \S{2}-hard even if the
theory has either one or two extensions.

\end{corollary}

\begin{corollary}

Checking whether a default theory skeptically entails a
formula in the constrained or rational semantics is
\P{2}-hard even if the theory has either one or two
extensions.

\end{corollary}

Assume that a bijective faithful polynomial time translation
from constrained or rational default logic into Reiter or
justified default logic exists. By Lemma~\ref{one-or-two},
the validity of $\exists X \forall Y . F$ can be
translated in polynomial time into the question of whether
$\neg a \wedge b$ is equivalent to an extension of the
theory $\l D,W \r$ of Lemma~\ref{one-or-two}. This question
can in turn be translated into the question of whether the
translated theory has an extension equivalent to $\neg a
\wedge b \wedge G$ for some formula $G$ not mentioning the
variables in $\{a,b\} \cup X \cup Y$; this is indeed
required for this extension to be var-equivalent to $Cn(\neg
a \wedge b)$. On the other hand, the assumption that the
translation is bijective ensures that at most one such
formula $G$ exists. This means that the problem can be
solved in Reiter or justified default logic with an
unambiguous Turing machine. Formally, we have the following
result.

\begin{theorem}

If there exists a polynomial time bijective faithful
translation from either constrained or rational default
logic into Reiter or justified default logic then
$\np^\np \subseteq \up^\np$.

\end{theorem}

\proof The problem of whether $E$ is equivalent to a
constrained or rational extension of a default theory is
\S{2}-hard even if the theory is known to have only one
other extension which is inconsistent with $E$.

Assuming that a translation such in the statement of the
theorem exists, the problem can be translated into checking
the existence of a subset of the consequences of the
translated theory $E'$ such that $EE'$ is an extension of
the translated theory. On the other hand, if such an $E'$
exists is unique. Therefore, the test can be done by
unambiguously guessing a subset of the consequences $E'$ and
then checking whether $EE'$ is a Reiter or justified
extension of the translated theory; since the latter problem
can be solved with a polynomial number of calls to an \np\
oracle \cite{rosa-99,libe-extensions}, the whole problem
would be in $\up^\np$.~\qed

 %

 %

\section{Strongest-Extension Translations}

In this section, we show some bijective faithful reductions
that require polynomial time only once given one of the
strongest extensions $E$ of the original theory is known.
Such translations are polynomial-time given a formula that
is equivalent to $E$; since $E$ is deductive closure of the
consequences of some defaults in the theory, a formula of
polynomial size that is equivalent to $E$ exists. Since
these translations produce a polynomially sized result, they
are polynomial-size.

\subsection{From Rational to Constrained}

We show a reduction from rational to constrained default
logic. This translation is proved to work by relating each
rational process of the original theory with a constrained
process of the translated theory. In the sequel, we refer to
the rational process of the original theory as the
``simulated process'' and the constrained process of the
translated theory as the ``simulating process''. We also
refer to a formula $a \rightarrow \gamma$ as ``$\gamma$
conditioned to $a$''. A simulated process is related to its
simulating process as follows:

\begin{enumerate}

\item the consequences of the simulated process are derived
conditioned to $a$ (that is, prepended with $a \rightarrow$)
in the simulating process;

\item the justifications of the simulated process are derived
conditioned to $b$ (that is, prepended with $b \rightarrow$)
in the simulating process;

\item both consequences and justifications of the simulated
process are in the justifications of the simulating process
using a different alphabet.

\end{enumerate}

Formally, every process $\Pi$ is related to its simulating
process $\Pi'$ as follows:\eatpar

\begin{eqnarray*}
\cons(\Pi') &=&
\{a \rightarrow \gamma ~|~ \gamma \in \cons(\Pi)\} \cup \\
&&
\{b \rightarrow \beta ~|~ \gamma \in \just(\Pi)\} \\
\\
\just(\Pi') &=&
\{\gamma[X/X'] ~|~ \gamma \in \cons(\Pi)\} \cup \\
&&
\{\beta[X/X'] ~|~ \gamma \in \just(\Pi)\}
\end{eqnarray*}

Let $\l D,W \r$ be the original default theory, where
$D=\{d_1,\ldots,d_m\}$. The background theory $W$ is
translated into $a \rightarrow W$. This means that $[]$ and
its simulated process satisfy the relationship above. We now
define the defaults of the translated theory in such a way
the above relationship remains satisfied.

For every default $d_i = \frac{\alpha:\beta}{\gamma}$ of the
original theory, we have two defaults which correspond to
different conditions of applicability of the original
default in the original theory in the rational semantics.

\begin{description}

\mathitem[
\frac
{a \rightarrow \alpha:\beta[X/X'] \wedge \gamma[X/X']}
{z_i (a \rightarrow \gamma) (b \rightarrow \beta)}
]%

The precondition of this default is $a \rightarrow \alpha$
is entailed in the simulating process if and only if the
precondition of the original default is entailed by the
simulated process.

The justification of this default is consistent if and only
if the original default is applicable and does not produce a
failure in the simulated default. Indeed, its justification
is consistent with $\just(\Pi')$ if and only if the set of
all consequences and justifications of the simulated process
is consistent with $\gamma \wedge \beta$.

Finally, the consequence of this default satisfies the
condition that the consequences and justifications of the
original default are added to the consequences of the
simulating process with the assumptions $a$ and $b$.

\mathitem[
\frac{(a \rightarrow \alpha)(ab \rightarrow \neg \beta):}
{z_i}
]%

This default is only applicable when the precondition of the
original default is entailed but its justification is
inconsistent with the current set of justifications and
consequences. This is because all justifications and
consequences of the simulated process can be obtained by
assuming both $a$ and $b$. Therefore, if assuming both $a$
and $b$ then $\beta$ is false, then we are in the situation
in which the original default cannot be applied because of
its justification.

\end{description}

These two defaults can be applied only if the precondition
of the original default is entailed. Moreover, the first
default can be applied if the simulated default can be
applied without generating a failure. The second default can
be applied only if the simulated default cannot be applied.
Therefore, if $a \rightarrow \alpha$ is true, $z_i$ can be
produced unless the simulated default can be applied but
produces a failure. The idea is that we generate $a$
whenever we are in the condition in which all applicable
defaults are applied but a failure is not generated. This is
obtained by the following default.

\[
\frac
{\bigwedge_{d_i \in D} ((a \rightarrow \prec(d_i)) \rightarrow z_i):a \neg E}
{a \neg b z_1 \ldots z_m}
\]

This default can be applied only if, for each default whose
precondition is entailed ($(a \rightarrow \prec(d_i)$ is
true), either the default cannot be applied or it can be
applied without generating a failure ($z_i$ is true).
Therefore, this default can be applied only if the simulated
process is successful and closed (the justification $\neg E$
is explained below.)

The consequence of this default include $a$, thus making all
consequences that have been derived so far unconditioned. It
also includes $\neg b z_1 \ldots z_m$; this formula entails
all formulae conditioned to $b$ and all formulae $z_i$
derived so far. This way, the resulting extension does not
depend on which formulae $b \rightarrow \beta$ and $z_i$
have already been generated.

In order for the theory to simulate the original one we have
to generate $E$ whenever the simulated process ends up in a
failure. To simplify the matter, we allow $E$ to be
generated at any time unless the above default has been
applied. In the other way around, if we can arrive to a
point in which the above default is applied, we do not
generate $E$; in all other cases $E$ can be generated.

\[
\frac{:}
{\neg a \neg b E z_1 \ldots z_m}
\]

The consequence of this default include $\neg a$, $\neg b$,
and $z_1 \ldots z_m$, which entail all consequences of the
defaults that have already been applied. This way, the
generated extension does not depend on which other defaults
have been applied.

The only case in which this default cannot be applied is
when $a$ has been already derived. Such a derivation cannot
be accomplished by the defaults above because we always
check the consistency of a formula $\gamma$ before
generating $a \rightarrow \gamma$. Therefore, this default
is blocked only if $a$ has not be generated by the previous
default.

We can therefore conclude that, if the simulated process if
successful and closed we can apply the previous default and
generate the corresponding extension. In all other cases,
this last default is applicable, and generates the known
extension $E$.

The precondition $\neg E$ of the second-last default is used
to avoid $E$ to be generated both by this and the last
previous default. In order for this to work, we require $E$
to be an extension such that $E' \models E$ does not hold
for any other extension $E'$. In this case, the
inconsistency of $E$ and the extension to be different from
$E$ are the same condition. As a result, if generating $a$
produces an extension that is different from $E$, then the
default cannot be applied because the generated extension
would be inconsistent with its justification.

The rationale of the proof can be therefore summarized as follows:

\begin{enumerate}

\item the consequences of the simulated theory are drawn
conditioned to $a$, and preconditions are checked
conditioned to $a$;

\item the justifications of the simulated theory are drawn
conditioned to $b$;

\item both consequences and justifications of the original
theory are in the justifications of this theory but
rewritten in another alphabet;

\item we can always generate $\neg a \neg b E$;

\item whenever all defaults that can be applied are applied,
and that does not result in a failure, we generate $a \neg
b$, making all consequences unconditioned and all
justifications void;

\item we generate $a \neg b$ only when the produced
extension is different from $E$.

\end{enumerate}

\proofnewpage
\subsubsection{Formal Definition of the Translation}

Let us now formally prove the correspondence between each
theory and its translation. We assume that $\l D,W \r$,
where $D=\{d_1,\ldots,d_m\}$, is a theory that has
extensions and that $E$ is one of its strongest extensions,
that is, for no other extension $E'$ it holds $E' \models
E$. We prove some claims relating the processes of the
original and translated theory. First, we define the
following two functions.

\begin{eqnarray*}
T^e_{RC}\left(\frac{\alpha:\beta}{\gamma},~i\right)
&=&
\frac
{a \rightarrow \alpha:\beta[X/X'] \wedge \gamma[X/X']}
{z_i (a \rightarrow \gamma) (b \rightarrow \beta)}
\\
T^n_{RC}\left(\frac{\alpha:\beta}{\gamma},~i\right)
&=&
\frac{(a \rightarrow \alpha)(ab \rightarrow \neg \beta):}{z_i}
\end{eqnarray*}

The translated default theory is obtained by translating
each default separately to two ones and then adding the
following two further defaults to it.

\begin{eqnarray*}
T^g_{RC}(\{d_1,\ldots,d_m\})
&=&
\frac{\bigwedge_{d_i \in D} ((a \rightarrow \prec(d_i))
\rightarrow
z_i):a \neg E}
{a \neg b z_1 \ldots z_m}
\\
T^s_{RC}(\{d_1,\ldots,d_m\})
&=&
\frac{:}
{\neg a \neg b E z_1 \ldots z_m}
\end{eqnarray*}

The translation is defined as follows.

\begin{eqnarray*}
T_{RC}(\l \{d_1,\ldots,d_m\},W \r)
&=& \l T_{RC}(D),T_{RC}(W) \r
\\
\mbox{where}
\\
T_{RC}(D)
&=&
\{
  T^e_{RC}(d_i,i),
  T^n_{RC}(d_i,i) 
     ~|~ 1 \leq i \leq m
\}
\cup
\\
&&
\{
  T^g_{RC}(\{d_1,\ldots,d_m\}) ,~
  T^s_{RC}(\{d_1,\ldots,d_m\})
\}
\\
T_{RC}(W)
&=&
(a \rightarrow W) \wedge W[X/X']
\end{eqnarray*}

\proofnewpage
\subsubsection{Preliminary Results}

In this section we show some general properties of
propositional entailment. In particular, we consider
formulae in the form $a \rightarrow A$, that is, formulae
that are ``conditioned'' to a given variable $a$. The
translation uses formulae conditioned to variables in the
background theory and in the preconditions and consequences
of some defaults.

\begin{lemma}
\label{conditioned}

For any triple of formulae $A$, $B$, and $C$ not containing
the variables $a$ and $b$, it holds $A \models C$ if and
only if $(a \rightarrow A) \wedge (b \rightarrow B) \models
a \rightarrow C$.

\end{lemma}

\proof We first assume to the contrary, that $A \models C$
but $(a \rightarrow A) \wedge (b \rightarrow B) \not\models
(a \rightarrow C)$. The latter implies that  $(a \rightarrow
A) \wedge (b \rightarrow B) \wedge \neg (a \rightarrow C)$
is satisfiable. This formula is equivalent to the following
other ones:

\begin{eqnarray*}
(a \rightarrow A) \wedge (b
\rightarrow B) \wedge \neg (a \rightarrow C)
& \equiv &
(a \rightarrow A) \wedge (b
\rightarrow B) \wedge \neg (\neg a \vee C) \\
& \equiv &
(a \rightarrow A) \wedge (b
\rightarrow B) \wedge a \wedge \neg C \\
& \equiv &
A \wedge (b \rightarrow B) \wedge a \wedge \neg C
\end{eqnarray*}

If this formula is consistent, there exists a model that
satisfies both $A$ and $\neg C$, thus violating the
assumption that $A \models C$.

Let us now prove the converse. We assume that $(a
\rightarrow A) \wedge (b \rightarrow B) \models (a
\rightarrow C)$ but $A \not\models C$. The latter condition
implies that there exists a model $M$ that satisfies both
$A$ and $\neg C$. Let us consider the model $M'$ that
extends $M$ by the assignment $a = \true$ and $b = \false$.
This model satisfies $a$, $A$, $\neg C$, and $b \rightarrow
B$. As a result, it satisfies $A \wedge (b \rightarrow B)
\wedge a \wedge \neg C$, which we proved to be equivalent to
$(a \rightarrow A) \wedge (b \rightarrow B) \wedge \neg (a
\rightarrow C)$. As a result, $(a \rightarrow A) \wedge (b
\rightarrow B) \not\models  a \rightarrow C$, contrary to
the assumption.~\qed

The second lemma is about conditioning with two variables.

\begin{lemma}
\label{d-conditioned}

For any triple of formulae $A$, $B$, and $C$ not containing
the variables $a$ and $b$, it holds $AB \models C$ if and
only if $(a \rightarrow A) \wedge (b \rightarrow B) \models
(ab \rightarrow C)$.

\end{lemma}

\proof Assume that $AB \models C$. We have that $ab AB$ is
equivalent to $ab (a \rightarrow A)(b \rightarrow B)$. Since
$AB$ entails $C$, we have that $ab (a \rightarrow A)(b
\rightarrow B) \models C$. This condition can be rewritten
as $(a \rightarrow A)(b \rightarrow B) \models (ab
\rightarrow C)$.

Let us now prove the converse. Assume that $(a \rightarrow
A) \wedge (b \rightarrow B) \models (ab \rightarrow C)$ but
$AB \not\models C$. Then, we have a model that satisfies
$AB$ and $\neg C$ at the same time. By adding the assignment
of $a=\true$ and $b=\true$ we obtain a model that satisfies
$a$, $b$, $A$, $B$, and $\neg C$. This formula therefore
satisfies $(a \rightarrow A) \wedge (b \rightarrow B)$ but
does not satisfy $ab \rightarrow C$, contrary to the
assumption.~\qed

In the following, we use the above lemmas together with the
following property.

\begin{property}
\label{noshare}

If $K$ is satisfiable and does not share any variables with
$A$ and $C$, then $A \models C$ if and only if $K \wedge A
\models C$.

\end{property}

\proofnewpage
\subsubsection{e-Sequences}

We show a correspondence between each sequence of defaults
$\Pi$ of the original theory and the following sequence of
defaults of the translated theory:

\[
T^e_{RC}([d_{i_1},\ldots,d_{i_k}])
=
[T^e_{RC}(d_{i_1},i_1),\ldots,T^e_{RC}(d_{i_k},i_k)]
\]

The consequences of $\Pi$ and $T^e_{RC}(\Pi)$ are related as
follows:

\begin{eqnarray}
\lefteqn{T_{RC}(W) \wedge \cons(T^e_{RC}(\Pi)) \equiv}
\nonumber \\
& \equiv &
(a \rightarrow W) \wedge W[X/X'] \wedge \cons(T^e_{RC}(\Pi))
\nonumber
\\
& \equiv &
(a \rightarrow W) \wedge W[X/X'] \wedge
\bigcup_{d_j \in \Pi}
\Big(
  z_j \wedge
  (a \rightarrow \cons(d_j)) \wedge
  (b \rightarrow \just(d_j))
\Big)
\nonumber
\\
& \equiv &
W[X/X'] \wedge
\Big( \bigcup_{d_j \in \Pi} z_j \Big)
\wedge
(a \rightarrow (W \wedge \cons(\Pi))) \wedge
(b \rightarrow \just(\Pi))
\label{consequences}
\end{eqnarray}

Entailment of the precondition of a default from a sequence
$\Pi$ corresponds to the same condition on the translated
theory and sequence, as the following lemma shows.

\begin{lemma}
\label{preconditions}

It holds $W \cup \cons(\Pi) \models \prec(d_i)$ if and only
if $T_{RC}(W) \wedge \cons(T^e_{RC}(\Pi)) \models
\prec(T^e_{RC}(d_i,i))$.

\end{lemma}

\proof By the above correspondence between the conclusions
of $\Pi$ and of $T^e_{RC}(\Pi)$, the condition $T_{RC}(W)
\wedge \cons(T^e_{RC}(\Pi)) \models \prec(T^e_{RC}(d_i,i))$
can be rewritten as follows:

\[
W[X/X'] \wedge
\Big( \bigcup_{d_j \in \Pi} z_j \Big)
\wedge
(a \rightarrow (W \wedge \cons(\Pi))) \wedge
(b \rightarrow \just(\Pi))
\models a \rightarrow \prec(d_i)
\]

Since $W[X/X']$ and $\cup z_i$ do not share variables with
the other formulae, by Property~\ref{noshare} this condition
is equivalent to:

\[
(a \rightarrow (W \wedge \cons(\Pi))) \wedge
(b \rightarrow \just(\Pi))
\models a \rightarrow \prec(d_i)
\]

By Lemma~\ref{conditioned}, this condition is equivalent to
$W \cup \cons(\Pi) \models \prec(d_i)$.~\qed

This lemma proves that the precondition of a default $d_i$
is entailed after the application of $\Pi$ in the original
theory if and only if the same condition holds for the
translated theory and defaults. The following result is an
immediate consequence of this lemma.

\begin{corollary}
\label{processes}

A sequence of defaults $\Pi$ is a process of $\l D,W \r$ if
and only if $T^e_{RC}(\Pi)$ is a process of $T_{RC}(\l D,W
\r)$.

\end{corollary}

The justifications of $\Pi$ and of $T^e_{RC}(\Pi)$ are
related in a similar way. In particular, we can show the
following equivalence about global consistency.

\begin{eqnarray}
\lefteqn{T_{RC}(W) \cup
\just(T^e_{RC}(\Pi)) \cup
\cons(T^e_{RC}(\Pi))} \nonumber \\
& \equiv &
(a \rightarrow W) \wedge W[X/X'] \cup
\just(T^e_{RC}(\Pi)) \cup
\cons(T^e_{RC}(\Pi)) \nonumber \\
& \equiv &
(a \rightarrow W) \wedge W[X/X'] \cup \nonumber \\
&&
\bigcup_{d_i \in D} \just(d)[X/X'] \wedge \cons(d)[X/X'] \cup
\nonumber \\
&&
\bigcup_{d_i \in \Pi} z_i (b \rightarrow \just(d)) (a \rightarrow \cons(d))
\nonumber \\
& \equiv &
\bigcup_{d_i \in \Pi} z_i \cup
\nonumber \\
&&
W[X/X'] \cup \just(\Pi)[X/X'] \cup \cons(\Pi)[X/X'] \cup
\nonumber \\
&&
(a \rightarrow W) \cup (a \rightarrow \cons(\Pi)) \cup
(b \rightarrow \just(\Pi))
\nonumber \\
& \equiv &
\bigcup_{d_i \in \Pi} z_i \cup
\nonumber \\
&&
(W \cup \just(\Pi) \cup \cons(\Pi))[X/X'] \cup
\nonumber \\
&&
(a \rightarrow (W \cup \cons(\Pi)) \cup
(b \rightarrow \just(\Pi))
\label{justifications}
\end{eqnarray}

The consistency of this formula is easy to relate to the
corresponding formula of $\Pi$.

\begin{lemma}
\label{success}

A sequence of defaults $\Pi$ is a globally successful
process of $\l D,W \r$ if and only if $T^e_{RC}(\Pi)$ is a
globally successful process of $T_{RC}(\l D,W \r)$.

\end{lemma}

\proof Formula \ref{justifications} is consistent if and
only if $W \cup \just(\Pi) \cup \cons(\Pi)$ is consistent.
Indeed, Formula \ref{justifications} contains $W \cup
\just(\Pi) \cup \cons(\Pi)$ rewritten on a new alphabet plus
other formulae that are always satisfiable by setting both
$a$ and $b$ to $\false$.~\qed

This property can be pushed a little further by showing that
if $T^e_{RC}(\Pi)$ is a globally successful process, then
its justifications and consequences are not only consistent
with the background theory but also with any other set of
variables from $Z \cup \{a\}$.

\begin{lemma}
\label{addingAZ}

The process $T^e_{RC}(\Pi)$ is globally successful if and
only if $T_{RC}(W) \cup \cons(T^e_{RC}(\Pi)) \cup
\just(T^e_{RC}(\Pi)) \cup Z \cup \{a,\neg b\}$ is
consistent.

\end{lemma}

\proof If $T^e_{RC}(\Pi)$ is globally successful, by
Equation~(\ref{justifications}) $(W \cup \just(\Pi) \cup
\cons(\Pi))[X/X']$ is consistent, which means that $W \cup
\just(\Pi) \cup \cons(\Pi)$ is also consistent, that is, it
has a model. By adding the assignment $a=\true$, $b=\false$,
and $z_i=\true$, we obtain a model of $T_{RC}(W) \cup
\cons(T^e_{RC}(\Pi) \cup \just(T^e_{RC}(\Pi)) \cup Z \cup
\{a,\neg b\}$.~\qed

This lemma adds something to the previous one: not only
$\Pi$ is globally successful if and only if $T^e_{RC}(\Pi)$
is globally successful, but the addition of other defaults
having a subset of $Z \cup \{a,\neg b\}$ as consequences is
irrelevant to the successfulness of $T^e_{RC}(\Pi)$.

Having proved that processes correspond to processes and
successful processes correspond to successful processes,
what remains to be proved is that closed processes
correspond to closed processes. This is however not true in
the presented reduction, which is based on the idea of
treating specially those processes for which a default is
globally applicable but would lead to global inconsistency.

\proofnewpage
\subsubsection{n-Sequences}

For any sequence of defaults $\Pi$ of $\l D,W \r$, we
consider the following sequence of defaults of $T_{RC}(\l
D,W \r)$. In this formula, $\prod$ denotes a sequence of
elements. The following is not necessarily a process, nor
$\Pi$ has been assume to be a process.

\[
T^n_{RC}(\Pi)
=
T^e_{RC}(\Pi) \cdot
\prod_{
W \cup \cons(\Pi) \models \prec(d_i)
\atop
W \cup \just(\Pi) \cup \cons(\Pi) \cup \just(d_i) \models \bot
}
T^n_{RC}(d_i,i)
\]

The following lemma is about the preconditions of the
defaults $T^n_{RC}(d_i,i)$ that are in a sequence
$T^n_{RC}(\Pi)$.

\begin{lemma}
\label{global}

If $T^n_{RC}(d_i,i)$ is a default in $T^n_{RC}(\Pi)$, then
$T_{RC}(W) \cup \cons(T^e_{RC}(\Pi)) \models
\prec(T^n_{RC}(d_i,i))$ if and only if $d_i$ is not globally
applicable to $\Pi$.

\end{lemma}

\proof The assumption that $T^n_{RC}(d_i,i) \in
T^n_{RC}(\Pi)$ implies that $W \cup \cons(\Pi) \models
\prec(d_i)$. Therefore, $d_i$ is globally applicable to
$\Pi$ if and only if $W \cup \cons(\Pi) \cup \just(\Pi) \cup
\just(d_i)$ is consistent.

The precondition of $T^n_{RC}(d_i,i)$ is a conjunction of
the precondition of $T^e_{RC}(d_i,i)$ and $ab \rightarrow
\neg \just(d_i)$. By Lemma~\ref{preconditions}, the
precondition of $T^e_{RC}(\Pi)$ is entailed by
$T^e_{RC}(\Pi)$ if and only if $W \cup \cons(\Pi) \cup
\prec(d_i)$, which is true by assumption.

Regarding the second condition, the entailment of $ab
\rightarrow \neg \just(d_i)$ from $T^e_{RC}(\Pi)$ in
formulae is:

\[
W[X/X'] \wedge
\Big( \bigcup_{d_j \in \Pi} z_j \Big) \wedge
(a \rightarrow (W \wedge \cons(\Pi))) \wedge
(b \rightarrow \just(\Pi))
\models ab \rightarrow \neg \just(d_i)
\]

The formulae $W[X/X']$ and $\cup z_i$ can be neglected by
Property~\ref{noshare} because they do not share variables
with the other formulae. By Lemma~\ref{d-conditioned}, the
resulting condition $(a \rightarrow (W \wedge \cons(\Pi)))
\wedge (b \rightarrow \just(\Pi)) \models ab \rightarrow
\neg \just(d_i)$ is equivalent to $W \cup \cons(\Pi) \cup
\just(\Pi) \models \neg \just(d_i)$, which is the opposite
of the global applicability of $d_i$ to $\Pi$ because $W
\cup \cons(\Pi) \models \prec(d_i)$ holds by
assumption.~\qed

The fact that the defaults $T^n_{RC}(d_i,i)$ have no
justifications and a very simple consequence has the effect
that their order in $T^n_{RC}(\Pi)$ does not matter.

\begin{lemma}
\label{nposition}

For any sequence $\Pi$, the sequence $T^n_{RC}(\Pi)$ is a
globally successful process if and only if $T^e_{RC}(\Pi)$
is a globally successful process and $T_{RC}(W) \cup
\cons(T^e_{RC}(\Pi)) \models \prec(T^n_{RC}(d_i,i))$ for
every $T^n_{RC}(d_i,i) \in T^n_{RC}(\Pi)$.

\end{lemma}

\proof The defaults $T^n_{RC}(d_i,i)$ do not have
justifications, and their consequences are contained in $Z$.
As a result, the set of justifications and consequences of
$T^n_{RC}(\Pi)$ is exactly the same as that of its first
part $T^e_{RC}(\Pi)$ with a subset of $Z$ added to it. By
Lemma~\ref{addingAZ}, this set is consistent if and only if
$T^e_{RC}(\Pi)$ is globally successful.

Regarding these sequences being processes or not, the
consequence of a defaults $T^n_{RC}(d_i,i)$ does not affect
the precondition of another default of the same kind
$T^n_{RC}(d_j,j)$. Therefore, two defaults of this kind can
always be swapped. As a result, if $T^n_{RC}(\Pi)$ is a
process then any of its defaults $T^n_{RC}(d_i,i)$ can be
moved to be immediately after $T^e_{RC}(\Pi)$. This proves
that $T^e_{RC}(\Pi) \cdot [T^n_{RC}(d_i,i)]$ must be a
process, which is the same as $T_{RC}(W) \cup
\cons(T^e_{RC}(\Pi)) \models \prec(T^n_{RC}(d_i,i)$ because
the default $T^n_{RC}(d_i,i)$ has no justification. The
converse is true by the monotonicity of the underlying
logic.~\qed

These two lemmas can be condensed as follows.

\begin{corollary}

For any sequence $\Pi$, the sequence $T^n_{RC}(\Pi)$ is a
globally successful process if and only if $T^e_{RC}(\Pi)$
is a globally successful process and, for any $d_i$ such
that $T^n_{RC}(d_i,i) \in T^n_{RC}(\Pi)$, it holds that
$d_i$ is not globally applicable to $\Pi$.

\end{corollary}

The condition of $T^n_{RC}(\Pi)$ being a globally successful
process can be linked to $\Pi$ being a rational process.

\begin{lemma}
\label{rational}

The sequence $\Pi$ is a rational process of $\l D,W \r$ if
and only if $T^n_{RC}(\Pi)$ is a globally successful process
of $T_{RC}(\l D,W \r)$.

\end{lemma}

\proof The sequence $\Pi$ is a rational process if and only
if $\Pi$ is globally successful and every default not in
$\Pi$ is not globally applicable to $\Pi$.

The global success of $\Pi$ is equivalent to the global
success of $T^e_{RC}(\Pi)$ by Lemma~\ref{success}. We
therefore only have to prove that every $d_i \not\in \Pi$ is
not globally applicable to $\Pi$ if and only if, for every
$T^n_{RC}(d_i,i) \in T^n_{RC}(\Pi)$, it holds that
$T^e_{RC}(\Pi) \cdot [T^n_{RC}(d_i,i)]$ is a process. On the
other hand, the above corollary proves exactly this
claim.~\qed

\proofnewpage
\subsubsection{Permutation of Defaults}

The correspondence between the processes of the original and
the translated theory is not bijective. Indeed, many
processes of the translated theory generate the extension
$E$, while the same extension can be generated by one or few
processes in the original theory. One reason is that more
than one constrained process might generate an extension
that is var-equivalent to $E$. On the other hand, we can
prove that all such processes generate the same extension.

\begin{lemma}
\label{s-not-g}

All constrained processes of $T_{RC}(\l D,W \r)$ containing
$T^s_{RC}(D)$ generate the extension $W[X/X'] \neg a \neg b
E z_1 \ldots z_m$.

\end{lemma}

\proof The formula $W[X/X'] \neg a \neg b E z_1 \ldots z_m$
is the conjunction of the background theory and the
conclusion of $T^s_{RC}(D)$. If a process contains this
default, its generated extension contains this formula. We
therefore only have to prove that the generated extension
does not include other formulae that are not entailed by
this one.

Let $\Pi$ be a rational process of $T_{RC}(\l D,W
\r)$ that contains $T^s_{RC}(D)$. Since this process is
successful and this default has $\neg a$ as a conclusion,
the process does not contain $T^e_{RC}(D)$, which contains
$a$ as a precondition. All other defaults in $T_{RC}(\l D,W
\r)$ have consequences that are entailed by $\neg a \neg b E
z_1 \ldots z_m$; therefore, their presence in the process
does not affect the generated extension.~\qed

This lemma shows that all processes containing $T^s_{RC}(D)$
generate the same extension, which is var-equivalent to $E$.
Therefore, we can exclude these processes and the extension
$E$ from consideration. In other words, we have to prove a
bijection between the extensions of the two theories besides
the extension $E$ and $W[X/X'] \neg a \neg b E z_1 \ldots
z_m$. What we actually prove is that there is a bijection
between processes modulo permutation on the order of the
defaults.

\begin{lemma}
\label{g-not-s}

A constrained processes of $T_{RC}(\l D,W \r)$ not
containing $T^s_{RC}(D)$ contains $T^g_{RC}(D)$, and
therefore does not generate an extension that is
var-equivalent to $E$.

\end{lemma}

\proof The default $T^s_{RC}(D)$ has no precondition and no
justification. It is therefore applicable to every process,
provided that its consequence is not inconsistent with the
conclusions and justifications of the other defaults in the
process. On the other hand, the consequence of $T^s_{RC}(D)$
is consistent with all justifications and conclusions of all
defaults but $T^g_{RC}(D)$. Therefore, if $T^s_{RC}(D)$ is
not in a process, this process must include $T^g_{RC}(D)$.
Since $\neg E$ is a justification of this default, the
generated extension cannot be var-equivalent to $E$.~\qed

We have therefore divided the constrained processes of
$T_{RC}(\l D,W \r)$ into two groups: those containing
$T^s_{RC}(D)$ and generating the extension $W[X/X'] \neg a
\neg b E z_1 \ldots z_m$ and those including $T^g_{RC}(D)$
and generating an extension that is not var-equivalent to
$E$. The consequences of the translated defaults are all in
the form $a \rightarrow \gamma$ and $b \rightarrow \beta$.

We now prove that processes can be put in a normal form in
which defaults $T^e_{RC}(d,i)$ occur first. We first prove
that these defaults can always be put before defaults
$T^n_{RC}(d,i)$.

\begin{lemma}
\label{eposition}

If $T_{RC}(\l D,W \r)$ has a globally successful process in
which a default $T^e_{RC}(d,i)$ follows a default
$T^n_{RC}(d,i)$, this default theory also has a globally
successful process in which the two defaults are swapped.

\end{lemma}

\proof The only condition that makes swapping two
consecutive defaults in a process impossible is when the
precondition of the second default is not entailed without
the consequence of the first. It is easy to show that this
is not the case in the assumption of the lemma.

Indeed, the consequence of $T^n_{RC}(d_j,j)$ is $z_j$. The
background theory does not contain $z_j$, while the
conclusions of all other defaults either do not contain
$z_j$ or are in the form $z_j \wedge A$, for some formula
$A$. Since the precondition of $T^e_{RC}(d_i,i)$ does not
contain $z_j$, Property~\ref{noshare} proves that this
precondition is entailed from the previous defaults if and
only if it is entailed by the previous defaults minus
$T^n_{RC}(d_j,j)$.~\qed

We now prove that a default $T^e_{RC}(d,i)$ cannot follow
the default $T^g_{RC}(D)$.

\begin{lemma}
\label{noafter}

No constrained process of $T_{RC}(\l D,W \r)$ contains
$T^g_{RC}(D)$ followed by $T^e_{RC}(d,i)$.

\end{lemma}

\proof Consider the first default $T^e_{RC}(d,i)$ that
follows $T^g_{RC}(D)$. All defaults between these two are in
the form $T^n_{RC}(d,i)$ because this process does not
contain $T^s_{RC}(D)$ and $T^e_{RC}(d,i)$ is the first one
after $T^g_{RC}(D)$. By Lemma~\ref{eposition}, the default
$T^e_{RC}(d,i)$ can be moved immediately after the default
$T^g_{RC}(D)$. In other words, if there exists a globally
successful process in which $T^e_{RC}(d,i)$ follows
$T^g_{RC}(D)$, then the following is also a globally
successful process:

\[
\Pi = \Pi_1 \cdot [T^g_{RC}(D),T^e_{RC}(d_i,i)] \cdot \Pi_2
\]

This is a process. Therefore, the precondition of
$T^e_{RC}(d_i,i)$ is entailed by $\Pi_1 \cdot
[T^g_{RC}(D)]$, which can be rewritten as:

\begin{eqnarray*}
\lefteqn{
T_{RC}(W) \cup
\cons(\Pi_1) \cup \cons(T^g_{RC}(D)) \models
\prec(T^e_{RC}(d_i,i))
} \\
& \mbox{ iff } &
(a \rightarrow W) \wedge W[X/X'] \cup
\bigcup_{T^e_{RC}(d_j,j) \in \Pi_1} \cons(T^e_{RC}(d_j,j))
\cup \\
&&
\bigcup_{T^n_{RC}(d_j,j) \in \Pi_1} \cons(T^n_{RC}(d_j,j)) \cup
\{a, \neg b, z_1, \ldots, z_m\} \models
a \rightarrow \prec(d_i) \\
& \mbox{ iff } &
(a \rightarrow W) \wedge W[X/X'] \cup
\bigcup_{T^e_{RC}(d_j,j) \in \Pi_1}
z_j(a \rightarrow \cons(d_j)) (b \rightarrow \just(d_j)) \cup
\\
&&
\Big( \bigcup_{T^n_{RC}(d_j,j) \in \Pi_1} z_j \Big)\cup
\{a, \neg b, z_1, \ldots, z_m\} \models
a \rightarrow \prec(d_i) \\
&& \mbox{since $a$ and $b$ are true and false, respectively} \\
& \mbox{ iff } &
W \wedge W[X/X'] \cup
\bigcup_{T^e_{RC}(d_j,j) \in \Pi_1}
z_j\cons(d_j)
\cup
\\
&&
\bigcup_{T^n_{RC}(d_j,j) \in \Pi_1} z_j \cup
\{a, \neg b, z_1, \ldots, z_m\} \models
\prec(d_i) \\
&& \mbox{removing subformulae according to Property~\ref{noshare}} \\
& \mbox{ iff } &
W \cup
\bigcup_{T^e_{RC}(d_j,j) \in \Pi_1} \cons(d_j) \cup
\{a\} \models
a \rightarrow \prec(d_i) \\
& \mbox{ iff } &
W \cup \bigcup_{T^e_{RC}(d_j,j) \in \Pi_1} \cons(d_j)
\models \prec(d_i) \\
&& \mbox{by Lemma~\ref{conditioned}} \\
& \mbox{ iff } &
(a \rightarrow W) \cup
\bigcup_{T^e_{RC}(d_j,j) \in \Pi_1} (a \rightarrow \cons(d_j))
\models a \rightarrow \prec(d_i)
\end{eqnarray*}

The formula preceeding $\models$ is a subformula of
$T_{RC}(W) \cup \cons(\Pi_1)$. As a result, $T_{RC}(W) \cup
\cons(\Pi_1) \models a \rightarrow \prec(d_i)$.

By assumption, $T^g_{RC}(D)$ is applied after $\Pi_1$.
Therefore, all its preconditions must be entailed at this
point. In particular, $(a \rightarrow \prec(d_i))
\rightarrow z_i$ must be entailed. Since $a \rightarrow
\prec(d_i)$ is true after $\Pi_1$, then $z_i$ must be true
as well. Since $T^e_{RC}(d_i,i)$ is not in $\Pi_1$ by
assumption, the only remaining default having $d_i$ as a
consequence is $T^n_{RC}(d_i,i)$. Therefore, we have that
$\Pi_1$ contains $T^n_{RC}(d_i,i)$.

A consequence of this fact is that the precondition of
$T^n_{RC}(d_i,i)$ is entailed from $\Pi_1$. Let us focus on
the second part of the precondition:

\begin{eqnarray*}
\lefteqn{T_{RC}(W) \cup \cons(\Pi_1)
\models ab \rightarrow \neg \just(d_i)} \\
& \mbox{ iff } &
(a \rightarrow W) \wedge W[X/X'] \cup \\
&&
\bigcup_{T^e_{RC}(d_j,j) \in \Pi_1} \cons(T^e_{RC}(d_j,j))
\cup
\bigcup_{T^n_{RC}(d_j,j) \in \Pi_1} \cons(T^n_{RC}(d_j,j)) \\
&&
\models ab \rightarrow \neg \just(d_i) \\
& \mbox{ iff } &
(a \rightarrow W) \wedge W[X/X'] \cup \\
&&
\bigcup_{T^e_{RC}(d_j,j) \in \Pi_1}
  z_j(a \rightarrow \cons(d_j))(b \rightarrow \just(d_j)) \\
&& \cup
\bigcup_{T^n_{RC}(d_j,j) \in \Pi_1} z_j \\
&&
\models ab \rightarrow \neg \just(d_i) \\
&& \mbox{removing the irrelevant parts by Property~\ref{noshare}} \\
& \mbox{ iff } &
(a \rightarrow W) \cup 
\bigcup_{T^e_{RC}(d_j,j) \in \Pi_1}
  (a \rightarrow \cons(d_j))(b \rightarrow \just(d_j))
\models ab \rightarrow \neg \just(d_i) \\
&& \mbox{by Lemma~\ref{d-conditioned}} \\& \mbox{ iff } &
W \cup 
\bigcup_{T^e_{RC}(d_j,j) \in \Pi_1}
  \cons(d_j) \wedge \just(d_j)
\models \neg \just(d_i) \\
&& \mbox{replacing $X$ with $X'$ everywhere} \\
& \mbox{ iff } &
W[X/X'] \cup \just(\Pi_1) \cup \cons(\Pi_1)
\models \neg \just(d_i)[X/X']
\end{eqnarray*}

The latter formula implies that $W[X/X'] \cup \just(\Pi_1)
\cup \cons(\Pi_1) \cup \just(T^e_{RC}(d_i,i)$ is
inconsistent. As a result, the process $\Pi$ is not globally
successful, contradicting the assumption.~\qed

The latter two lemmas, together, implies that every
constrained process of $T_{RC}(\l D,W \r)$ can be put in a
sort of ``normal form''.

\begin{corollary}
\label{normal}

For each globally successful process of $T_{RC}(\l D,W \r)$
containing $T^g_{RC}(D)$, there exists another successful
process that is composed of the same defaults, but all
defaults $T^e_{RC}(d_i,i)$ came first, followed by some
defaults $T^n_{RC}(d_i,i)$, followed by $T^g_{RC}(D)$
followed by some other defaults $T^n_{RC}(d_i,i)$.

\end{corollary}

\proofnewpage
\subsubsection{Correspondence of Extensions}

The correspondence between the rational processes of the
original theory and the constrained processes of the
translated theory is obtained as follows. For each sequence
of defaults $\Pi$ of the original theory, we consider the
following sequence of the translated theory.

\begin{eqnarray}
T_{RC}(\Pi)
&=&
T^n_{RC}(\Pi) \cdot [T^g_{RC}(D)]
\label{all-process}
\end{eqnarray}

We establish the following correspondence: each rational
process $\Pi$ of $\l D,W \r$ not having $E$ as an extension
corresponds to the constrained process $T_{RC}(\Pi)$ of
$T_{RC}(\l D,W \r)$, and vice versa. The converse is true in
the sense that for every constrained process of $T_{RC}(\l
D,W \r)$ there is an equivalent constrained process in which
all defaults are in the form of $T_{RC}(\Pi)$.

\begin{lemma}

If $T_{RC}(\Pi)$ is a globally successful process, then
there exists a sequence $\Pi'$ such that $T_{RC}(\Pi) \cdot
\Pi'$ is a constrained process of $T_{RC}(\l D,W \r)$ and
for all such $\Pi'$ it holds $\cons(T_{RC}(\Pi)) \models
\cons(\Pi')$.

\end{lemma}

\proof If $T_{RC}(\Pi)$ is a globally successful process, it
can be a non-constrained process only because it is not
maximal. On the other hand, the only applicable defaults are
in the form $T^n_{RC}(d_i,i)$ because of Lemma~\ref{g-not-s}
and Lemma~\ref{noafter}. The consequences of these defaults
are entailed by that of $T^g_{RC}(D)$.~\qed

\begin{lemma}

A formula $E'$ that is not var-equivalent to $E$ is an
extension of $T_{RC}(\l D,W \r)$ if and only if $E' =
T_{RC}(W) \cup \cons(T_{RC}(\Pi))$ and $T_{RC}(\Pi)$ is a
globally successful process.

\end{lemma}

\proof If $T_{RC}(\Pi)$ is a globally successful process,
then it can be completed to form a constrained process by
adding to it some defaults whose consequences are already
entailed by $T_{RC}(\Pi)$. Since $T_{RC}(\Pi)$ contains the
default $T^g_{RC}(D)$, which has $\neg E$ as a
justification, the generated extension is not $E$.

Let us assume that $E$ is a constrained extension of
$T_{RC}(\l D,W \r)$ that is not equivalent to $E'$. By
Lemma~\ref{s-not-g} and Lemma~\ref{g-not-s}, its generating
process contains $T^g_{RC}(D)$. By Corollary~\ref{normal},
the default theory contains a process with the same defaults
in which the defaults $T^e_{RC}(d_i,i)$ preceed all other
ones, followed by some defaults $T^n_{RC}(d_i,i)$ followed
by $T^g_{RC}(D)$ followed by some other defaults. Denoting
by $\Pi$ the set of defaults $d_i$ such that either
$T^e_{RC}(d_i,i)$ or $T^n_{RC}(d_i,i)$ is before
$T^g_{RC}(D)$ in this process, we have that this process can
be rewritten as $T_{RC}(\Pi) \cdot \Pi'$. Since this is a
constrained process, $T_{RC}(\Pi)$ is globally
successful.~\qed

The following lemma relates the rational process of the
original theory with the processes obtained by the function
$T_{RC}$.

\begin{lemma}

$\Pi$ is a rational process of $\l D,W \r$ not generating
$E$ as an extension if and only if $T_{RC}(\Pi)$ is a
globally successful process of $T_{RC}(\l D,W \r)$.

\end{lemma}

\proof By Lemma~\ref{rational}, $\Pi$ is a rational process
if and only if $T^n_{RC}(\Pi)$ is globally successful. Since
$T_{RC}(\Pi) = T^n_{RC}(\Pi) \cdot [T^g_{RC}(D)]$, if this
process if globally successful then $T^n_{RC}(\Pi)$ is
globally successful as well. Therefore, we only have to
prove that, if $\Pi$ is a rational process, then
$T^n_{RC}(\Pi) \cdot [T^g_{RC}(D)]$ is globally successful.
In particular, since $T^n_{RC}(\Pi)$ is globally successful
and remains so even if their consequences are added $\{a,
\neg b\} \cup Z$ by Lemma~\ref{addingAZ}, what remains to be
proved is only that the precondition of $T^g_{RC}(D)$ is
entailed from the process $T^n_{RC}(\Pi)$.

Since $\Pi$ is a rational process, for any default $d_i$
such that $W \cup \cons(\Pi) \models \prec(d_i)$ it holds
that either $d_i \in \Pi$ or that $W \cup \cons(\Pi) \cup
\just(\Pi) \models \neg \just(d_i)$. These three conditions
can be rephrased in the translated theory as follows.

\begin{description}

\item[$W \cup \cons(\Pi) \models \prec(d_i)$\rm .] This is
equivalent to $T_{RC}(W) \cup \cons(T^n_{RC}(\Pi)) \models a
\rightarrow \prec(d_i)$ by Lemma~\ref{preconditions};

\item[$d_i \in \Pi$\rm .] This means that $T^e_{RC}(d_i,i)
\in T^n_{RC}(\Pi)$, and therefore that $T_{RC}(W) \cup
\cons(T^n_{RC}(\Pi)) \models z_i$;

\item[$W \cup \cons(\Pi) \cup \just(\Pi) \models \neg
\just(d_i)$\rm .] This means that $T^n_{RC}(d_i,i) \in
T^n_{RC}(\Pi)$, and therefore that $T_{RC}(W) \cup
\cons(T^n_{RC}(\Pi)) \models z_i$.

\end{description}

As a result, since $\Pi$ is a rational process then, for
every index $i$ such that $T_{RC}(W) \cup
\cons(T^n_{RC}(\Pi)) \models a \rightarrow \prec(d_i)$ it
also holds that $T_{RC}(W) \cup \cons(T^n_{RC}(\Pi)) \models
z_i$. As a result, the precondition of $T^g_{RC}(D)$ is
entailed.~\qed

This lemma, together with Lemma~\ref{s-not-g} and
Lemma~\ref{g-not-s} allows proving the correctness of the
translation.

\begin{corollary}

For every rational process $\Pi$ of $\l D,W \r$ there are a
number of constrained processes of $T_{RC}(\l D,W \r)$ all
generating the same extension, which is var-equivalent to
the extension generated by $\Pi$, and vice versa.

\end{corollary}

 %

\subsection{From Reiter to Justified and Constrained}

In order to translate theories from Reiter to justified
default logic, we adopt a strategy slightly different from
the one used in the previous translation. Namely, we allow
the application of a default even if its justification is
violated; however, we do not then generate the extension in
this case (we generate the known extension instead). We
still replace $W$ with $a \rightarrow W$. Each default
$d_i=\frac{\alpha:\beta}{\gamma}$ is simulated by the two
following defaults:

\begin{description}

\mathitem[
\frac{a \rightarrow \alpha:\beta[X/X']}
{z_i (a \rightarrow \gamma)}
]%

This default is always applicable whenever the precondition
of the simulated default is entailed. In other words, the
justification of this default is always consistent at this
stage.

\mathitem[
\frac{(a \rightarrow \alpha)(a \rightarrow \neg \beta):}
{z_i}
]%

This default can only be applied if the precondition of the
original default is entailed but its justification is
inconsistent with the current set of consequences.

\end{description}

These defaults can only be applied if the precondition of
the original default is entailed. In particular, if the
justification of the original default is contradicted, we
have a choice of applying the first or the second default.
If the original default is instead applicable, we are forced
applying the first default. The fact that the first default
can be applied even if the original default cannot will not
be a problem, as these processes will be at a later time
forced to generate the known extension $E$.

As above, we have the default that generates the known
extension, and which can always be applied:

\[
\frac:{\neg a E z_1 \ldots z_n}
\]

This default can be applied provided that $a$ has not been
generated. On the converse, if the defaults that have been
applied correspond to a successful process, we can instead
generate $a$ and produce the extension,by applying the
following default:

\[
\frac{\bigwedge_{d_i \in D}
((a \rightarrow \prec(d_i)) \rightarrow z_i):\neg E}
{a (X \equiv X') z_1 \ldots z_m}
\]

Generating $a$ makes all consequences of the defaults that
have been applied unconditioned. On the other hand, $X
\equiv X'$ makes each formula $\beta[X/X']$ equivalent to
$\beta$. This is done to check the justifications of all
applied defaults of the first kind. If such a default has
been applied while its original default could not been
applied because of its justification, the addition of $X
\equiv X'$ would create a failure. This means that the last
default is not applied, as justified default logic does not
allow generating a failure. As a result, this last default
can only be applied if the simulated process is successful.
Otherwise, the only applicable default is the one producing
the known extension $E$.

The justification $\neg E$ forbids $E$ to be generated in
two different ways, and work only if $E$ is one of the
logically strongest extensions of the original theory.

Since justified default logic can be translated in
polynomial time into constrained default logic, it follows
that Reiter default logics can be translated into
constrained default logic given one of the strongest
extensions.

 %

 %

\section{Polysize Translations}

In this section, we show the effects of the existence of
some polynomial-size translations between variants of
default logic. Existing translations have been shown in the
previous sections: polynomial-time translations and
translations that work given a strongest extension are also
polysize translations. The following result shows the
ability of rational and constrained default logic to express
the consistency of a formula with a partial interpretation.

\begin{lemma}
\label{assignment}

For any formula $F$ over variables $X \cup Y \cup Z$ it is
possible to build in polynomial time a default theory $\l
D,W \r$ such that the following hold, where $F|_{\omega_Z}$
is the formula obtained from $F$ by replacing each variable
in $Z$ with its truth value in $\omega_Z$:

\begin{enumerate}

\item for every truth assignment $\omega_Z$ on the variables
$Z$, the formula $\omega_Z \neg a \neg b$ is a rational and
constrained extension of $\l D,W \r$;

\item for every truth assignment $\omega_Z$ on the variables
$Z$, the formula  $\omega_Z \neg a b$ is a rational and
constrained extension of $\l D,W \r$ if and only if $\exists
X \forall Y . F|_{\omega_Z}$ is valid;

\item $\l D,W \r$ has no other rational or constrained
extension.

\end{enumerate}

\end{lemma}

\proof Let $X=\{x_1,\ldots,x_n\}$ and
$Z=\{z_1,\ldots,z_m\}$. The default theory that corresponds
to $F$ is $\l D,\emptyset \r$, where $D$ is defined as
follows; $a$, $b$, and $\{k_1,\ldots,k_m\}$ are new
variables.

\begin{eqnarray*}
D &=&
\left\{
\left.
\frac{:z_ik_i}{z_ik_i} ,~
\frac{:\neg z_ik_i}{\neg z_ik_i}
\right|
1 \leq i \leq m
\right\}
\cup
\\
&&
\left\{
\left.
\frac{K:x_i}{a \rightarrow x_i} ,~ 
\frac{K:\neg x_i}{a \rightarrow \neg x_i} 
\right|
1 \leq i \leq n
\right\}
\cup
\\
&&
\left\{
\frac{K(a \rightarrow F):\neg a b}{\neg a b} ,~
\frac{K:\neg a \neg b}{\neg a \neg b}
\right\}
\\
&& \mbox{ where } K=k_1 \wedge \cdots \wedge k_m
\end{eqnarray*}

This set of defaults require a choice on all variables $z_i$
to be taken before applying any other default. As a result,
every extension of this theory contains a complete truth
assignment over the variables $Z$.

Once such a truth assignment has been obtained, we can apply
the default $\frac{K:\neg a \neg b}{\neg a \neg b}$, thus
obtaining the extension of the point 1. of the statement.

The only way of blocking this default it to apply the second
last default. In turn, this default can be applied only if
some of the defaults of the second subset can be applied in
such a way the resulting conclusions $a \rightarrow
\omega_X$ entail $a \rightarrow F|_{\omega_Z}$ regardless of
the value of $Y$. Therefore, $\omega_Z \neg a b$ is an
extension if and only if $\exists X \forall Y .
F|_{\omega_Z}$ is valid.~\qed

The default theory of the proof does not produce the same
Reiter and justified extensions. This is because the
defaults $\frac{K:x_i}{a \rightarrow x_i}$ and $\frac{K:\neg
x_i}{a \rightarrow \neg x_i}$ can coexist in the same Reiter
or justified process without making it unsuccessful. To make
these defaults to contradict the justification of each other
one would need to change their justifications to $a \wedge
x_i$ and $a \wedge \neg x_i$, respectively; this however
would make the generation of $\neg a$ by the last two
defaults impossible.

This lemma is based on the ability of constrained and
rational default logics to collect the justifications of the
applied defaults without making them appear in the
conclusions. This is the reason why extension-checking is
harder in these two semantics than in Reiter and justified
default logics.

This idea constitutes the base of a possible proof of
non-existence of a bijective translation from rational or
constrained default logics to Reiter or justified default
logics. Namely, if such a translation existed, then one
would be able to solve the set of QBF problems $\exists X
\forall Y . F|_{\omega_Z}$ for every $\omega_Z$ by first
producing a rational or constrained default theory,
translating it to Reiter or justified default logic, and
then checking for the existence of an extension containing
$\omega_Z \neg a b$. For a fixed interpretation $\omega_Z$,
such a translation would be certainly feasible in a
polynomial amount of space. The point is that a bijective
faithful translation would need to produce a theory that has
one or two extensions for every interpretation $\omega_Z$
over the variables $Z$.

The problem with this line of proof is that, in the theory
that results from the translation, we cannot simply check
whether $\omega_Z \neg a b$ is an extension. Indeed, since
new variables are allowed, an extension $\omega_Z \neg a b$
is in general translated into an extension $\omega_Z \neg a
b G$, where $G$ is a formula built over the new variables
introduced by the translation.

For this reason, we consider the problem of checking whether
a formula is equivalent to part of an extension. This way,
we could check $\exists X \forall Y . F|_{\omega_Z}$ by
checking whether $\omega_Z \neg a b$ can be extended to form
an extension of the theory that results from the
translation. We can restrict to the case in which the theory
is known to have at most one extension extending the given
formula. Indeed, in the lemma above, only a single extension
containing $\omega_Z \neg a b$ may possibly exist; the
result of a bijective reduction is a single extension, if
any, containing $\omega_Z \neg a b$.

\

\def\ppS{\mbox{$\pp^{\Sigma^{\rm p}_2}$}}
\def\count{{\sf C}}

A majority Turing machine is a nondeterministic Turing
machine that output ``yes'' if and only if at least half of
the computation paths lead to acceptance. The class \pp\  is
the class of problems solved by a majority Turing machine
that works in polynomial time. Similarly, $\pp^A$ is the
class of problems solved by a majority Turing machine
working in polynomial time and equipped with an oracle that
solves the problem $A$ in constant time. The class $\pp^\C$,
where \C\  is a class of problems, is defined as the union
the classes of $\pp^A$ for every $A \in \C$.

A slightly different characterization of classes defined in
terms of oracles and nondeterministic Turing machines is in
terms of the counting quantifier $\count$. This quantifier
extends both $\exists$ and $\forall$ by allowing the minimal
number of assignments making a formula valid to be specified
arbitrarily. Wagner~\cite{wagn-86} and Toran~\cite{tora-91}
have shown that $\pp^{\rm K} = \count {\rm K}$ for every
class ${\rm K}$ that is defined in terms of quantifiers
$\count$, $\exists$, and $\forall$. Besides the superficial
difference on the bound, this result proves that an oracle
majority Turing machine can be restricted to make exactly
one call to the oracle in each path of computation without a
power loss.

\begin{theorem}

Deciding whether $|ext(T)| \geq k$ is $\count \forall
\exists \p$ complete for constrained and rational default
logic.

\end{theorem}

\proof The problem can be solved by counting the number of
processes that generate an extension. Since two processes
can generate the same extension, we define an ordering over
processes and only count the minimal one for each extension.
Given a default theory $T$, we define $\proc(T)$ to be its
selected processes and $minproc(T)$ its minimal selected
processes.

Given a default theory $\l D,W \r$, we add an arbitrary
linear ordering $<$ on the set of the defaults. A linear
ordering can be then defined over the processes: $\Pi <
\Pi'$ if and only if either $\Pi$ is shorter than $\Pi'$, or
$\Pi(i) < \Pi'(i)$ where $i$ is the first index for which
$\Pi(i) \not= \Pi'(i)$. Counting the extensions can be done
by counting the minimal processes:

\begin{eqnarray*}
\Pi \in minproc(T) &
\mbox{ iff } &
\Pi \in \proc(T) \\
&& \forall \Pi' ~.~
\Pi' \not\in \proc(T) ~\vee~
(\cons(\Pi') \not\equiv \cons(\Pi)) ~\vee~
\Pi < \Pi'
\end{eqnarray*}

A process $\Pi$ is in this set if and only if it is the
minimal process generating the extension $W \cup
\cons(\Pi)$. As a result, we can count the number of
extension of $T$ by counting the number of processes in
$minproc(T)$. Since deciding whether a process is in
$minproc(T)$ is in $\forall \exists \p$, deciding whether
their number is greater than a number $k$ is in $\count
\forall \exists \p$.

We prove the hardness of the problem by showing a reduction
from the problem of establishing whether the number of truth
assignments $\omega_Z$ over variables $Z$ such that a
formula $\exists X \forall Y . F|_{\omega_Z}$ is valid is
greater than or equal to a given bound. This problem is
$\count \exists \forall \p$ complete \cite{wagn-86}.

By Lemma~\ref{assignment}, the formula $\exists X \forall Y
. F|_{\omega_Z}$ is valid for $Z = \omega_Z$ exactly when
$\omega_Z \neg a b$ is an extension of the theory $\l D,W
\r$ of the lemma. Besides these extensions, the theory $\l
D,W \r$ has also exactly $2^{|Z|}$ extensions. Therefore,
checking whether the number of truth assignments $\omega_Z$
satisfying the condition above is greater than or equal to
$k$ is equivalent to checking whether $\l D,W \r$ has at
least $|ext(T)| \geq 2^{|Z|} + k$ extensions.~\qed

The same problem for Reiter and justified default logic is
slightly simpler.

\begin{theorem}

Checking whether $|ext(T)| \geq k$ is in $\count \exists$
for Reiter and justified default logic.

\end{theorem}

\proof We show that the problem is in $\pp^\np$, which is
equal to $\count \exists$. Checking whether a subset of
$\cons(D)$ is an extension of a theory for the considered
two semantics is in \D{2}: checking whether $E \in ext(T)$
can be solved by a polynomial number of calls to an \np\
oracle (actually, a logarithmic number suffices). Counting
the number of extensions of $T$ can be solved by counting
the number of nondeterministic paths of a Turing machine that
has one such path for every $D' \subseteq D$ and calls the
oracle for checking whether $E=\cons(D')$ is an extension.
Some nondeterministic paths have to be added to make the
bound $k$ to correspond exactly to one half of the
nondeterministic paths.~\qed

The complexity of the problem is therefore lower for Reiter
and justified default logics than for constrained and
rational default logics. Of course, this result is not
useful by itself, as the non-existence of a polynomial-time
translation is already established.

We denote by $T \leadsto T'$ the condition of existence of a
bijective faithful translation from $T$ to $T'$. This
condition can be formalized as follows, where $\equiv_T$
indicates var-equivalence over the variables of $T$.

\begin{eqnarray*}
T \leadsto T' & \mbox{iff} &
\forall E ~.~ E \in ext(T) \leftrightarrow
\exists E' ~.~ EE' \in ext(T') \\
&& \forall E' E'' ~.~ E' \in ext(T') ~ E'' \in ext(T') ~
(E' \equiv_T E'') \rightarrow E' = E''
\end{eqnarray*}

Checking the first line of the right-hand side of this
equation is in \P{3}\  because $EE' \in ext(T')$ is \D{2}\
and therefore in \S{2}: as a result, $\exists E' ~.~ EE' \in
ext(T')$ is in \S{2}\  as well. In the second check, the
dominating operation is to check the opposite of $E'
\equiv_T E''$, and checking var-equivalence is in \P{2}.

The idea is as follows: assume that, for every $T$, there
exists a theory $T'$ of polynomial size such that $T
\leadsto T'$. If this is the case, we can check the number
of extensions of $T$ by first guessing a theory $T'$ of
polynomial size, and then checking whether $T \leadsto T'$
and doing the check on the number of extensions on $T'$.
Formally:

\begin{eqnarray*}
|ext(T)| \geq k 
&\Leftrightarrow&
\exists T' ~.~
(T \leadsto T') ~\wedge~  |ext(T')| \geq k
\\
&\Leftrightarrow&
\forall T' ~.~
(T \leadsto T') ~\rightarrow~ (|ext(T')| \geq k)
\end{eqnarray*}

The first line reformulates the problem with an existential
quantifier $\exists$ and a formula that is in $\forall
\exists \forall \p$ and one that is in $\count \exists \p$.
The second line gives a similar result; note that $T
\leadsto T'$ is this time used in reverse because it is in
an antecedent of an implication.

The above conditions allow solving the problem in two
different ways, leading to membership to the following
inclusion:

\[
\count \exists \forall \p
\subseteq
\exists ( \forall\exists\forall \p \cup \count\exists \p) \cap
\forall ( \exists\forall\exists \p \cup \count\exists \p)
\]

Some results that hold for these classes are: every class
$\count K$ is closed under complementation \cite{tora-91},
and therefore $\count \exists K = \count \forall K$; both
$\exists K$ and $\forall K$ are included into $\count K$
\cite{tora-91}; and $PH \subseteq \p^{\count \p}$
\cite{toda-89}. Applying these results to the first class
above we get:

\[
\begin{array}{rcll}
\exists ( \forall\exists\forall \p \cup \count\exists \p)
& \subseteq &
\exists (\p^\pp \cup \count \forall \p)
& \mbox{Toda}
\\
& = &
\exists (\np^\pp \cup \count \forall \p)
& \mbox{because }\p \subseteq \np
\\
& = &
\exists (\np^{\count \p} \cup \count \forall \p)
& \mbox{because } \pp=\pp ^ \p \mbox{ and } \pp^{\rm K}=\count {\rm K}
\\
& = &
\exists (\exists \count \p \cup \count \forall \p)
& \mbox{since } \np^{\count {\rm K}} = \exists \count {\rm K}
\\
& \subseteq &
\exists (\exists \count \forall \p \cup \exists \count \forall \p)
& \mbox{adding quantifier can only enlarge classes}
\\
& = &
\exists \exists \count \forall \p
& \mbox{classes with $\exists$ in front are closed under union}
\\
&=&
\exists \count \forall \p
& \mbox{two quantifiers of the same type}
\end{array}
\]

For the second class, we obtain a similar result:

\begin{eqnarray*}
\forall ( \exists\forall\exists\p \cup \count\exists\p)
& \subseteq &
\forall (\p^\pp \cup \count\exists\p)
\\
& \subseteq &
\forall (\forall\count\p \cup \count\exists\p)
\\
& \subseteq &
\forall (\forall\count\exists\p \cup \forall\count\exists\p)
\\
& = &
\forall\count\exists\p
\end{eqnarray*}

Therefore, the assumption of existence of a bijective
faithful translation would imply that $\count \exists
\forall \p$ is contained in both $\exists\count\forall\p$
and $\forall\count\exists\p$. This condition can be restated
as: a counting quantifier can be swapped with either an
existential or a universal one.

If
$\count \exists \forall \p \subseteq \exists\count\forall\p$
then
$\exists \count \exists \forall \p
\subseteq
\exists\exists\count\forall\p
=
\exists\count\forall\p
\subseteq
\exists \count \exists \forall \p$,
and therefore
$\exists \count \exists \forall \p
=
\exists\count\forall\p$. With a similar proof one can
conclude that
$\forall \count \exists \forall \p
=
\forall \count \exists \p$.

 %

\section{Conclusions}

This article reports some results about the existence of
bijective-faithful translations among variants of default
logics. Translations between such variants have already been
investigated in the literature; some of such translations
are faithful: each extension of the original theory
corresponds to an equivalent extension of the translated
theory. This article makes the assumption that the
translations can introduce new variables; that implies that
faithful translations might not be bijective: each extension
of the original theory may correspond to many extensions of
the translated theory. We therefore considered translations
that are not only faithful but also create a bijection
between extensions.

The rationale of requiring such a bijection is that the
translated theory provides a more close simulation of the
original one. As an example, if one translates an instance
of the planning problem into Reiter default logic in such a
way each plan corresponds to an extension \cite{turn-97},
then translating this theory in another variant breaks this
correspondence if the translation is not bijective. If one
wants to enumerate all plans, and a non-bijective
translation has been applied, enumerating the extensions of
the translated theory does not automatically generate an
enumeration of all possible plans, because some plans may be
generated more than once. As an extreme example, a planning
instance having two plans $P_1$ and $P_2$ can be expressed
into a Reiter default theory having two extensions $E_1$ and
$E_2$. If one then converts this theory into constrained
default logic using a non-bijective translation, what may
result is a theory having a large number of extensions
corresponding to $E_1$ and a single one corresponding to
$E_2$. That means that enumerating all extensions of this
theory is likely to find a large number of extensions
corresponding to $P_1$ before finding the one corresponding
to $P_2$.

The same argument can be applied in general for the problem
of generating all extensions of a default theory, finding
the number of extensions, finding whether a theory has a
unique extension \cite{zhao-libe-02}, etc. All these
problems cannot be solved by first translating the theory
into a different semantics and then solving the problem in
that semantics, unless the translation is guaranteed to
translate every extension into a single extension.

 %

\appendix


\bibliographystyle{alpha}

\newpage


\end{document}